\documentclass[review]{elsarticle}
\usepackage[utf8]{inputenc}
\usepackage{array}
\usepackage{booktabs}
\usepackage{textcomp}
\usepackage{bm}
\usepackage{multirow}
\usepackage{amsmath}
\usepackage{amssymb}
\usepackage{subcaption}
\usepackage{miller}
\usepackage{gensymb}
\usepackage{float}

\newtheorem{thm}{Theorem}
\newdefinition{defn}{Definition}
\newproof{proof}{Proof}

\begin{document}

\title{Gradient-Based Training and Pruning of Radial Basis Function Networks with an Application in Materials Physics}

\author[cs,hiit]{Jussi Määttä}
\ead{jussi.maatta@alumni.helsinki.fi}
\author[cs,hiit]{Viacheslav Bazaliy}
\ead{viacheslav.bazaliy@alumni.helsinki.fi}
\author[phys]{Jyri Kimari\corref{cor1}}
\ead{jyri.kimari@helsinki.fi}
\author[phys]{Flyura Djurabekova}
\ead{flyura.djurabekova@helsinki.fi}
\author[phys]{Kai Nordlund}
\ead{kai.nordlund@helsinki.fi}
\author[cs,hiit]{Teemu Roos}
\ead{teemu.roos@cs.helsinki.fi}
\address[cs]{Department of Computer Science, University of Helsinki, Finland}
\address[hiit]{Helsinki Institute for Information Technology (HIIT), Helsinki, Finland}
\address[phys]{Helsinki Institute of Physics and Department of Physics, University of Helsinki, Finland}
\cortext[cor1]{Corresponding author}

\begin{abstract}
%
Many applications, especially in physics and other sciences, call for easily interpretable and robust machine learning techniques.
We propose a fully gradient-based technique for training radial basis function networks with an efficient and scalable open-source implementation.
We derive novel closed-form optimization criteria for pruning the models for continuous as well as binary data which arise in a challenging real-world material physics problem.
The pruned models are optimized to provide compact and interpretable versions of larger models based on informed assumptions about the data distribution.
Visualizations of the pruned models provide insight into the atomic configurations that determine atom-level migration processes in solid matter;
these results may inform future research on designing more suitable descriptors for use with machine learning algorithms.
\end{abstract}


\maketitle

\section{Introduction}

The \emph{radial basis function network} (RBFN) was introduced by
Broomhead and Lowe in 1988 \cite{Broomhead_init_rbf}. It is a simple
yet flexible regression model that can be interpreted as a feedforward
neural network with a single hidden layer. While it often cannot
compete in predictive accuracy against state-of-the-art black box
models such as deep neural networks or ensemble methods such as boosting, it is nevertheless an universal
approximator~\cite{univ_approx},
which means that its predictive accuracy can be improved to match any other predictor by increasing the amount of training data. Especially with large data sets, computational efficiency becomes an essential concern.

In this article, we describe a modern gradient-based RBFN implementation
based on the same computational machinery that is used in modern deep learning.
While gradient-based methods for training RBFN's have been criticized~\cite{OLS}
because of local optima, tailored training methods have been used
in the past with good results~\cite{sel_bp}.
We show that suitable optimizers, regularization techniques, and learning rate schedules
enable us to train large RBF networks
without overfitting and achieve predictive performance on~par with gradient-boosted decision trees.

Classical algorithms for training RBFN's usually follow a two-step
approach. In the first step, the parameter values describing centroid positions are determined,
for instance, by taking a random subset of input data points or applying
a suitable clustering algorithm. In the second step, the rest of the model parameters are computed using closed-form analytic
expressions \cite{intro_rbf,survey_old,survey_new}. Another approach
for finding RBFN parameters, the Orthogonal Least Squares (OLS) algorithm~\cite{OLS},
constructs the network sequentially by adding new centroids that maximize
the reduced variance of the error vector.

While these methods from late 1990's indeed produce reasonably well-per\-form\-ing
RBFN's, it is clear that the solutions thus found are suboptimal
because of the biases introduced by the multi-step algorithms.
More recent research~\cite{new_1,new_2} has mainly concentrated
on developing algorithms that aim to reach good predictive performance
while keeping the number of parameters in the network as low as possible~\cite{survey_new}.
With our method, it becomes straightforward to train an RBF network
with a large number of prototypes.

The parameters of an RBFN have easy-to-understand interpretations. For a network with tens of centroids,
this makes the RBFN as a whole quite interpretable; however, a network
with hundreds or thousands of centroids is in practice no more interpretable
than any large machine learning model.
Therefore, we also propose
a gradient-based RBFN pruning method that produces a smaller RBFN
that globally approximates the larger one. Our approach simultaneously
optimizes all parameters of the smaller RBFN and, crucially, minimizes
the expected discrepancy over the \emph{input data distribution},
not the input data points themselves. 
Hence a pruned RBF network thus obtained is optimized for an objective function
that maintains an explicit connection to the larger RBF network and
is therefore different from what one would use for directly training a small RBFN of the same size.

Interpretability is a great asset especially in those machine learning applications
where the learned patterns in the input data can give new valuable insight into
the mechanisms underlying the correlations between input and output. Data
sets where such patterns, sometimes too nuanced for human intuition to discover,
can be revealed by interpretable machine learning models are encountered e.g.
in natural sciences. In this paper, we apply RBFN's to a materials physics data
set that describes a subset of parameters for a kinetic Monte Carlo (KMC) model for surface diffusion
in copper~\cite{jyri2,jyri3}. In the KMC model, diffusion is interpreted as a
series of atomic migration events that have rates $\Gamma$
defined by their energy barriers $E_\mathrm{m}$:
\begin{equation}
\Gamma \propto \exp\left(\frac{-E_\mathrm{m}}{k_\mathrm{B}T}\right)
\end{equation}
where $k_\mathrm{B}$ is the Boltzmann constant and $T$ is temperature.
The barriers in turn are defined by the configuration of atoms around the
migrating atom. There are methods for computing the barriers
that correspond to different local atomic environments of the
event, but problems arise from the vast number of the environments.
Computing the barriers accurately is computationally expensive, so
either the accuracy or the number of different barriers has to be
compromised for parametrizing the KMC model. Machine learning
offers a way to interpolate and extrapolate barrier values based
on the incomplete data set of calculated barriers.

The input data assumes a perfect crystalline
structure, where variation only occurs in the occupation of fixed lattice
sites---either an atom sits at the site indexed~$i$, or not. Hence, the
input data can be losslessly converted to a binary sequence of occupation
numbers at lattice positions. The local atomic environment that is assumed
to affect the migration energy barrier is extended up to the second nearest
neighbors of the migrating atom. In the face-centered cubic copper lattice,
this comprises 26 lattice positions.

The data set does not span the entire 26-dimensional input space, that
in principle has $2^{26}\approx 67$\,million possible values; even
setting aside to computational cost of calculating such an immense
amount of migration energy barriers, there are other challenges related
to finding all of these barriers in crystalline surface systems~\cite{jyri1}.
In any case, it is necessary to interpolate or otherwise estimate
the barriers for the missing input values, and machine learning is
a promising approach to accomplish this. Were there a need to introduce
another element into the parameterization in addition to Cu, the input space
would suddenly grow to have $3^{26}$ possible values; likewise, expanding the local atomic
environment to include third nearest neighbors would grow the dimensionality
itself from 26 to~58. For input spaces like this, the only option is to use a method that is capable of generalizing from a small subset of all possible inputs.

The large RBFN's trained to the migration barrier data are then pruned,
and the input-associated weights are revealed to contain patterns that
correspond to physically meaningful three-dimensional symmetries, even though the
networks only ever saw the ``flat'' binary representations of the atomic environments.

We emphasize that our motivation for pruning is to achieve interpretability,
not to minimize the number of centroids used for making predictions.
Indeed, we will demonstrate with a materials physics data set
that one can train RBF networks with thousands of centroids without overfitting,
so there is no inherent need to use pruning as a form of regularization---and that
these large networks can be made interpretable with our proposed pruning method.
This approach is also fundamentally different from simply training a small RBF network:
limiting the number of centroids for the sake of interpretability would compromise predictive accuracy,
and in general the pruned networks we obtain are different from those one would get by directly training networks of the same size
because the objective functions are different.

Somewhat similar ideas on the the input data distribution and our
pruning criterion's functional form appear in the growing-and-pruning
(GAP) algorithm~\cite{GAP,GGAP} for constructing RBF networks. However,
GAP is designed for constructing RBFN's in sequential access cases
and does not consider pruning as a separate task.
Also related are two existing RBFN algorithms that incorporate pruning by
initially assigning each training data point its own centroid:
The early two-stage algorithm by Musavi et~al.~\cite{pruning_clust}
prunes the initial network by combining similar centroids using an unsupervised clustering approach,
then keeps the centroid locations fixed for the rest of the training process;
as discussed above, this is unlikely to converge to an optimal solution.
The more sophisticated algorithm of Leonardis et~al.~\cite{pruning_mdl}
alternates between gradient-based optimization steps (performed on the full data set)
and centroid removal steps based on the Minimum Description Length (MDL) principle;
unfortunately their approach does not seem to scale well to large data sets
because of the need to repeatedly solve a combinatorial optimization problem
with a search space exponential in the number of centroids.

The rest of this article is structured as follows. In Section~\ref{sec:framework},
we describe our gradient-based approach for training and pruning RBF
networks. In Section~\ref{sec:experiments}, we report experimental
results both on toy data and on a materials physics dataset. Finally,
we summarize our results and discuss future directions in Section~\ref{sec:end}.

\section{The RBFN Framework}

\label{sec:framework}

\subsection{Basic problem setting}

Suppose that we are given a data set $(\bm{x}_{i},y_{i})$, $i=1,2,\ldots,N$,
where the $\bm{x}_{i}\in\mathbb{R}^{D}$ are i.i.d.\ samples from
some distribution $p(\bm{x})$. Our task is to reconstruct an unknown
target function $g\colon\mathbb{R}^{D}\to\mathbb{R}$ with $y_{i}=g(\bm{x}_{i}) + \varepsilon_i$,
where the~$\varepsilon_i \sim N(0,\sigma^2)$ are i.i.d.\ noise with some fixed variance.

We model~$g$ with an RBF network~$f$, defined as follows:
\begin{defn}
\label{def:rbfn}A \emph{radial basis function network} (RBFN) is
a function $f\colon\mathbb{R}^{D}\to\mathbb{R}$ of the form
\begin{equation}
f\left(\mathbf{x}\right)=\alpha+\sum_{i=1}^{K}\beta_{i}\,\phi\left(\left\Vert \mathbf{x}-\bm{\Theta}_{i}\right\Vert \right)\label{eq:rbfn}
\end{equation}
where $\alpha\in\mathbb{R}$, $\bm{\beta}\in\mathbb{R}^{K}$, and
$\bm{\Theta}\in\mathbb{R}^{K\times D}$ are parameters,
$\phi$ is a kernel function,
and $\left\Vert {}\cdot{} \right\Vert$ denotes the Euclidean norm.
\end{defn}

We restrict ourselves to the Gaussian kernel $\phi(t)=e^{-\gamma t^{2}}$
with the parameter $\gamma>0$. The rows of $\bm{\Theta}$ in Definition~\ref{def:rbfn}
are called \emph{centroids} or \emph{prototypes} and can be interpreted
as pseudo--data points. An RBFN computes an input point's Euclidean
distance to each centroid, feeds these distances into the kernel,
and uses a weighted linear combination of the resulting values to
provide a prediction. Taken together, the weights and centroids define
areas of the input spaces with smaller or larger output values
and hence have natural interpretations.

We propose to fit an RBF network~$f$ to the observations by solving the optimization problem
\begin{equation}
\arg\min_{\gamma,\alpha,\bm{\beta},\bm{\Theta}}\left\{ \frac{1}{N}\sum_{i=1}^{N}\left[f(\bm{x}_{i}\mid\gamma,\alpha,\bm{\beta},\bm{\Theta})-y_{i}\right]^{2}\right\} \text{,}\label{eq:optimization_problem}
\end{equation}
i.e., by minimizing the mean squared error (MSE) between the~$f(\bm x_i)$ and the~$y_i$.
This is equivalent to finding a maximum likelihood solution for the probabilistic setting described above.

\subsection{Training}

To solve \eqref{eq:optimization_problem}, we use the same machinery
that is used in modern deep learning. More specifically, we provide
an open source implementation\footnote{Our implementation is available at \textbf{XXX} under an open source
license.} based on the \emph{PyTorch} framework~\cite{paszke2017automatic}
and the gradient-based \emph{Adam} optimizer~\cite{adam}. We use
$L_{2}$ regularization for all real-valued parameters (log-transformed
in the case of~$\gamma$) and train the RBFN over multiple epochs,
starting with random initialization and using minibatches. We use
early stopping with a validation set and use the validation set loss
to guide our learning rate schedule. Our approach is implemented as
a standard PyTorch module and is vectorized for efficiency.

\subsection{Pruning}

Assume that we have trained a large RBF network~$f_{K}$ with $K$~centroids.
Suppose then that we would like to find another RBFN~$f_{M}$ with
$M<K$ centroids that gives a good approximation to original RBFN.
Our principal motivation here is that a smaller network can be easier
to interpret. There are many ways in which one might specify what
is the best approximation to~$f_{K}$, but we propose to find an
RBFN that minimizes expected squared difference between the two RBFN's
predictions in the probability space of the data-generating process. 

Denote the parameters of the smaller RBFN by $(\gamma,\alpha,\bm{\beta},\bm{\Theta})$
and the (fixed) parameters of the original RBFN by $(k,a,\bm{b},\bm{Z})$.
Then the pruning task is to solve
\begin{equation}
\arg\min_{\gamma,\alpha,\bm{\beta},\bm{\Theta}}\left\{ \mathbb{E}_{p(\mathbf{x})}\left[\left(f_{K}(\mathbf{x})-f_{M}\left(\mathbf{x}\right)\right)^{2}\right]\right\} \label{eq:pruning_obj}
\end{equation}
where $p(\bm{x})$ is the data distribution.

By expanding the squares and using the linearity of expectation, we
can rewrite the expectation in~\eqref{eq:pruning_obj} as
\begin{multline}
\mathbb{E}_{p(\mathbf{x})}\left[\left(f_{K}(\mathbf{x})-f_{M}\left(\mathbf{x}\right)\right)^{2}\right]=\\
\begin{aligned} & (a-\alpha)^{2}\\
+\, & \sum_{i=1}^{K}\sum_{j=1}^{K}b_{i}\,b_{j}\,\mathbb{E}_{p(\mathbf{x})}\left[e^{-k\|\bm{x}-\bm{Z}_{i}\|^{2}-k\|\bm{x}-\bm{Z}_{j}\|^{2}}\right]\\
+\, & \sum_{i=1}^{M}\sum_{j=1}^{M}\beta_{i}\,\beta_{j}\,\mathbb{E}_{p(\mathbf{x})}\left[e^{-\gamma\|\bm{x}-\bm{\Theta}_{i}\|^{2}-\gamma\|\bm{x}-\bm{\Theta}_{j}\|^{2}}\right]\\
+\, & 2(a-\alpha)\left(\sum_{i=1}^{K}b_{i}\,\mathbb{E}_{p(\mathbf{x})}\left[e^{-k\|\bm{x}-\bm{Z}_{i}\|^{2}}\right]-\sum_{i=1}^{M}\beta_{i}\,\mathbb{E}_{p(\mathbf{x})}\left[e^{-\gamma\|\bm{x}-\bm{\Theta}_{i}\|^{2}}\right]\right)\\
-\, & 2\sum_{i=1}^{K}\sum_{j=1}^{M}b_{i}\,\beta_{j}\,\mathbb{E}_{p(\mathbf{x})}\left[e^{-k\|\bm{x}-\bm{Z}_{i}\|^{2}-\gamma\|\bm{x}-\bm{\Theta}_{j}\|^{2}}\right]\text{.}
\end{aligned}
\label{eq:e_decomposition}
\end{multline}
From the decomposition~\eqref{eq:e_decomposition} we see that the
optimization problem~\eqref{eq:pruning_obj} is reduced to the computation
of expectations of the form $\mathbb{E}_{p(\bm{x})}\left[e^{-k\|\bm{x}-\bm{u}\|^{2}-r\|\bm{x}-\bm{v}\|^{2}}\right]$.
The following theorem gives closed-form expressions for these expectations
under three practically relevant scenarios.
\begin{thm}
\label{thm:pruning_expectations}Let $k,r\geq0$ and $\bm{u},\bm{v}\in\mathbb{R}^{D}$.
\end{thm}

\begin{enumerate}
\item If $\bm{x}$ follows the distribution given by the mixture density
\[
p(\bm{x})=\prod_{i=1}^{D}\sum_{j=1}^{L_{i}}w_{ij}\:f_{N}(x_{i}\mid\mu_{ij},\sigma_{ij}^{2})
\]
where $f_{N}(\cdot\mid\mu,\sigma^{2})$ is the Gaussian density with
mean~$\mu$ and variance~$\sigma^{2}$, $w_{ij}\geq0$, and $\sum_{j}w_{ij}=1$
for all~$i$, then 
\begin{multline*}
\mathbb{E}_{p(\bm{x})}\left[e^{-k\|\bm{x}-\bm{u}\|^{2}-r\|\bm{x}-\bm{v}\|^{2}}\right]=\\
\prod_{i=1}^{D}\sum_{j=1}^{L_{i}}\frac{w_{ij}}{\sqrt{1+2\sigma_{ij}^{2}(k+r)}}\exp\left(-\frac{k(u_{i}-\mu_{ij})^{2}+r(v_{i}-\mu_{ij})^{2}+2\sigma_{ij}^{2}kr(u_{i}-v_{i})^{2}}{1+2\sigma_{ij}^{2}(k+r)}\right)\text{.}
\end{multline*}
\item If $\bm{x}$ follows the distribution given by
\[
p(\bm{x})=\prod_{i=1}^{D}f_{U}(x_{i}\mid a_{i},b_{i})
\]
where $f_{U}(\cdot\mid a,b)$ is the uniform density on the interval
$(a,b)$, then
\begin{multline*}
\mathbb{E}_{p(\bm{x})}\left[e^{-k\|\bm{x}-\bm{u}\|^{2}-r\|\bm{x}-\bm{v}\|^{2}}\right]=\\
\begin{aligned}\left(\frac{\pi}{4(k+r)}\right)^{D/2}\prod_{i=1}^{D}\, & \frac{1}{b_{i}-a_{i}}\,\exp\left(-\frac{kr(u_{i}-v_{i})^{2}}{k+r}\right)\\
 & \times\left[\textrm{erf}\left(\frac{k(b_{i}-u_{i})+r(b_{i}-v_{i})}{\sqrt{k+r}}\right)-\textrm{erf}\left(\frac{k(a_{i}-u_{i})+r(a_{i}-v_{i})}{\sqrt{k+r}}\right)\right]\text{.}
\end{aligned}
\end{multline*}
\item If $\bm{x}$ follows the distribution given by
\[
p(\bm{x})=\prod_{i=1}^{D}p_{\textrm{Bernoulli}}(x_{i}\mid q_{i})
\]
where $p_{\textrm{Bernoulli}}(\cdot\mid q)$ is the Bernoulli point
probability function for the outcomes~$\pm1$, then
\[
\mathbb{E}_{p(\mathbf{x})}\left[e^{-k\|\bm{x}-\bm{u}\|^{2}-r\|\bm{x}-\bm{v}\|^{2}}\right]=e^{-k\|\bm{u}+\bm{1}\|^{2}-r\|\bm{v}+\bm{1}\|^{2}}\prod_{i=1}^{D}\left[1+q_{i}\left(e^{4\left(ku_{i}+rv_{i}\right)}-1\right)\right]\text{.}
\]
\end{enumerate}
\begin{proof}
The proof is postponed to \ref{sec:proof}.
\end{proof}
Gaussian mixtures and the uniform distribution provide good approximations
to the majority of practical situations where the input data is free
of outliers. The binary case is relevant for more specific situations;
it arises, for instance, in modeling atomic environments in physical
simulations~\cite{jyri1,jyri2}.

Given the closed-form expressions for the expectations from Theorem~\ref{thm:pruning_expectations},
we can directly compute the pruning objective~\eqref{eq:e_decomposition}
for a given set of RBFN parameters. Each individual expectation has
a computational complexity of $O(D)$, which implies that~\eqref{eq:e_decomposition}
can be computed with $O(K^{2}D)$ operations, or if we omit constant
terms, $O(KMD)$. This is particularly remarkable in the Bernoulli
case, where a naïve computation of the expectation would involve summation
over $2^{D}$~terms.

In our PyTorch-based implementation, we provide efficient vectorized
implementations of the pruning objectives corresponding to the settings
where each input dimension has the distribution $N(0,1)$, $U(a,b)$,
or $\textrm{Bernoulli}(0.5)$. As in RBFN training, we use the \emph{Adam}
optimizer~\cite{adam} to minimize the pruning objective. We initialize
the smaller RBFN's centroids by sampling without replacement from
the larger RBFN, and as the objective is not convex, we use multiple
restarts to improve the quality of the final solution.

\section{Results and discussion}

\label{sec:experiments}

In all the experiments described in this section, we use the same
fixed hyperparameters when training RBF networks. Namely, we use a
batch size of~64 and weight decay ($L_{2}$ regularization parameter)~$10^{-5}$.
We use validation set--based early stopping with the following learning
rate schedule: We start with the learning rate~$10^{-2}$, and multiply
it by~$0.1$ when ten successive epochs have produced no improvement
for the validation set MSE. After each learning rate reduction, we
allow ten epochs before resuming the monitoring, and we stop training
when the learning rate goes below~$10^{-4}$.

When pruning RBF networks, we use a similar learning rate schedule,
but we start from~$10^{-3}$, and we stop when the objective has
not improved for ten iterations with the learning rate~$10^{-5}$. 

\subsection{Toy example with normal and uniform pruning}

First, we illustrate training and pruning with a simple toy data set.
We sample $x_{i}$, $i=1,2,\ldots,1000$, uniformly at random from
the interval $[-4,4]$ and compute $y_{i}=e^{-x_{i}^{2}}+0.2\cos(4x)$.
We then train an RBFN with 100~centroids, using 20\% of the data
points for early stopping. Afterwards, we prune the resulting RBFN
down to three centroids. For pruning, we use both the $N(0,1)$
and $U(-4,4)$ distributions, and for each choice, we select the best
pruned RBFN out of ten random initializations.

\begin{figure}
\centering{}\includegraphics[width=1\textwidth]{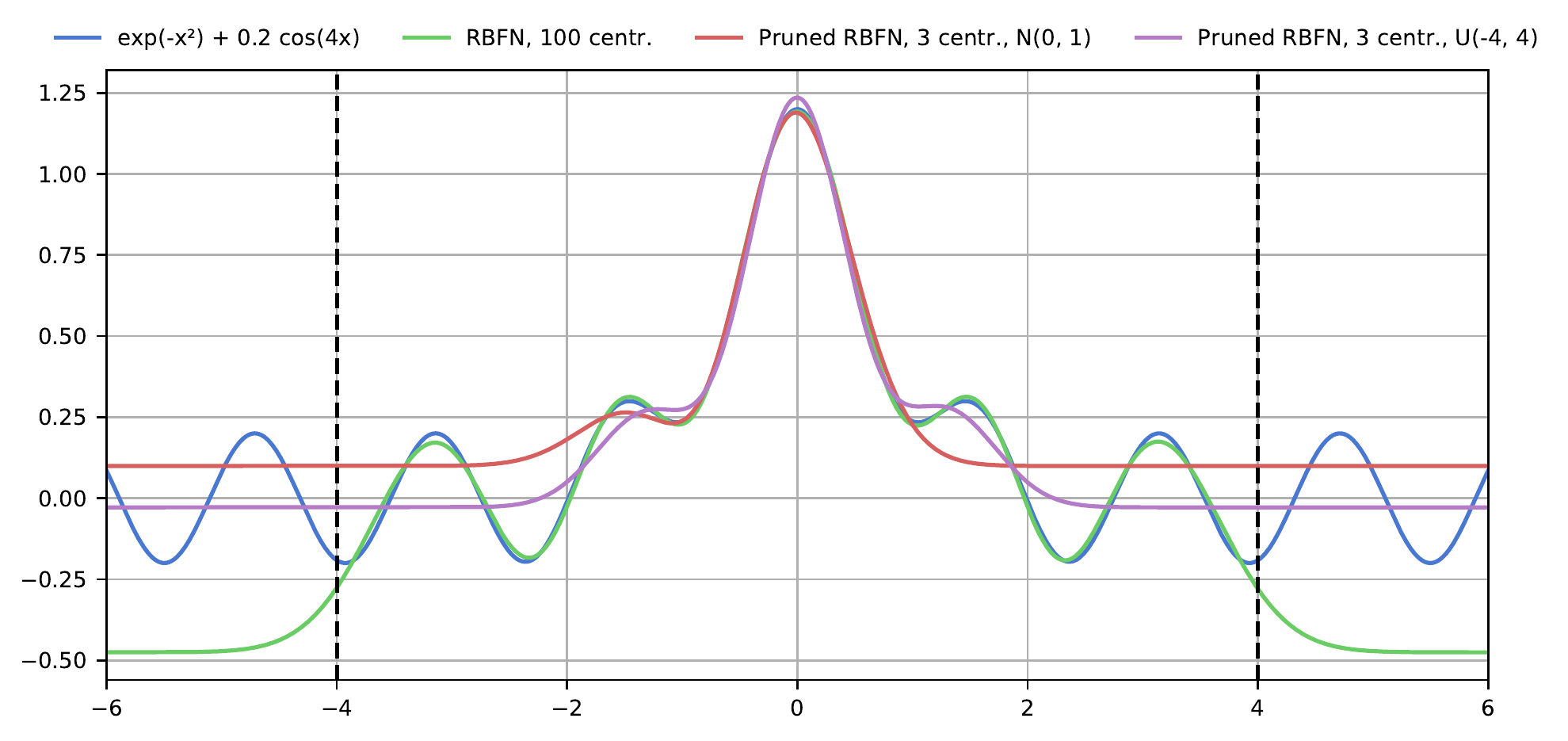}\caption{The effect of different pruning distributions.
The vertical lines mark the domain of the data set.}
\label{fig:toy}
\end{figure}

The results are shown in Figure~\ref{fig:toy}. The 100-centroid
RBFN has learned a very good approximation of the target function
within the data set and degenerates to a constant value elsewhere
as implied by Definition~\ref{def:rbfn}.
Roughly 95\% of the probability mass of $N(0,1)$ lies within the
interval $[-2,2]$, and within this interval the corresponding pruned
RBFN provides a good fit, but outside that interval the RBFN degenerates
to a constant with a suboptimal value; this is to be expected, as
pruning with this distribution should give little weight to values
outside the interval. The uniform-pruned RBFN also matches the central
peak well and converges to a more reasonable constant in the tails;
this focus on the tail area appears to be the cause for the slightly
worse approximation at around $x=\pm1$. 

\subsection{Migration energy barrier data}

\subsubsection{Predictive performance}

\label{subsec:rbfn_vs_xgb}

We evaluate the performance of gradient-based RBFN training on a data
set of migration energy barriers for copper surfaces~\cite{jyri3}.
The data set is further divided into three subsets corresponding to
the \hkl{100}, \hkl{110}, and \hkl{111} surfaces with
$N_{100}=1\,139\,281$,
$N_{110}=1\,495\,159$, and
$N_{111}=9\,017\,645$
data points, respectively.
Each data point consists of 26~binary features
and a real-valued response. We always use $100\,000$~data points
for training, $10\,000$~data points for early stopping, and leave
the rest for the test set.

\begin{figure}
  \centering
  \begin{subfigure}{0.19\linewidth}
    \centering
    \includegraphics[width=\linewidth]{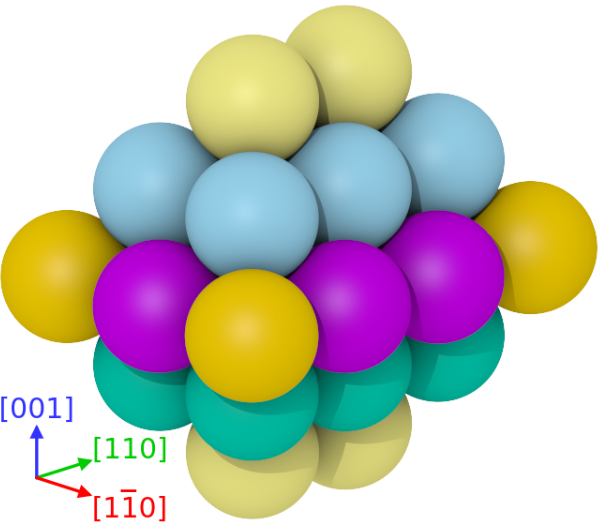}
  \end{subfigure}
  \begin{subfigure}{0.19\linewidth}
    \centering
    \includegraphics[width=\linewidth]{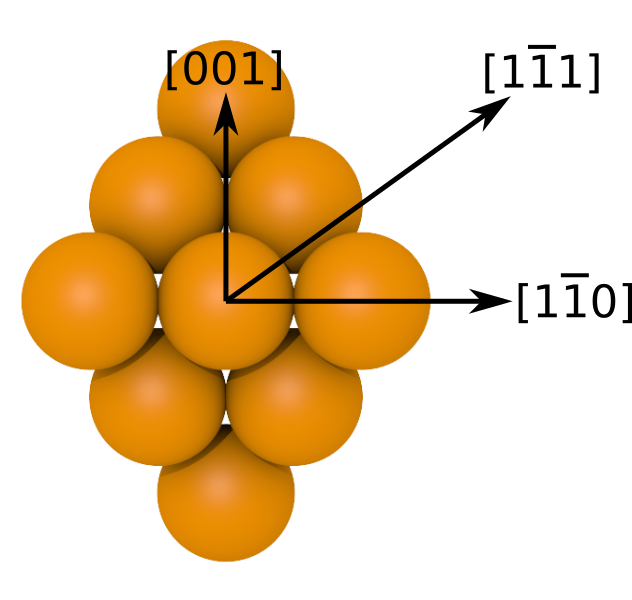}
  \end{subfigure}
  \begin{subfigure}{0.19\linewidth}
    \centering
    \includegraphics[width=\linewidth]{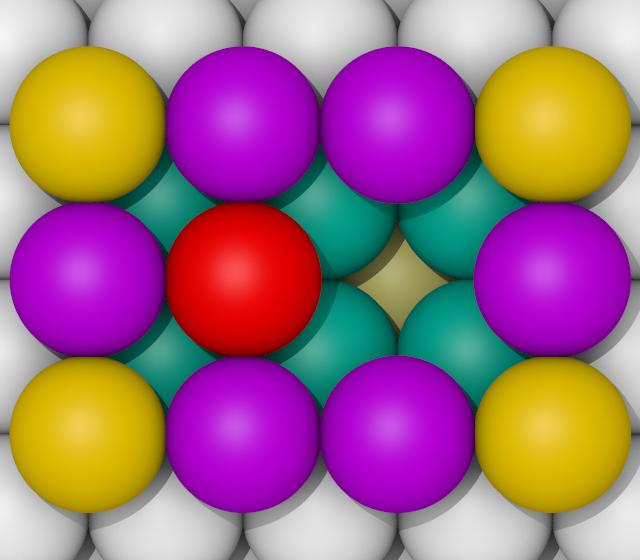}
  \end{subfigure}
  \begin{subfigure}{0.19\linewidth}
    \centering
    \includegraphics[width=\linewidth]{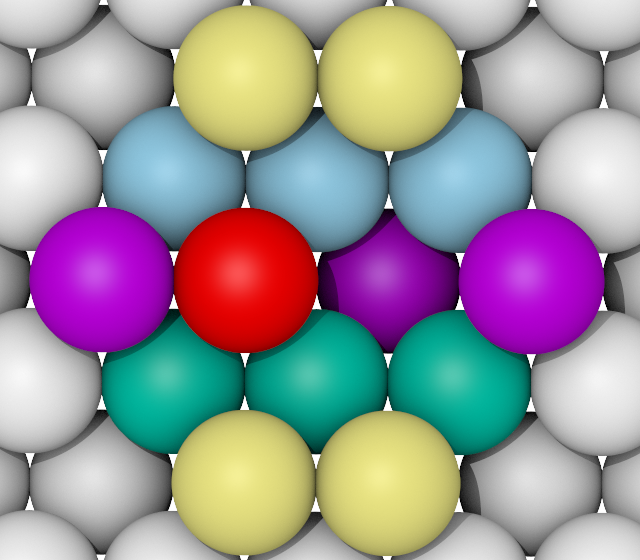}
  \end{subfigure}
  \begin{subfigure}{0.19\linewidth}
    \centering
    \includegraphics[width=\linewidth]{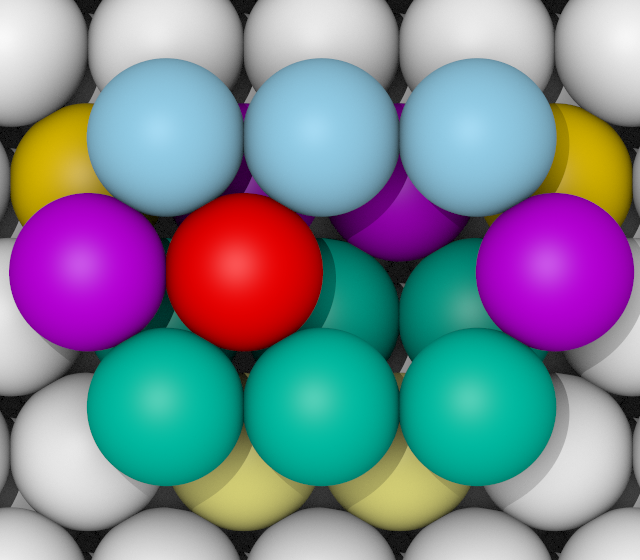}
  \end{subfigure}
  \caption{The 26 lattice sites within the local atomic environment in the leftmost panel are colored according to
  	   the atomic layer, and their distance from the migrating atom. The second
  	   panel shows examples of the \hkl<100>, the \hkl<110> and the
  	   \hkl<111> orientations with respect to the local atomic environment. These
  	   orientations are perpendicular to the \hkl{100}, the \hkl{110} and the \hkl{111}
  	   surfaces, respectively. Panels 3--5 show the
           embedding of the atomic environment in, from left to right, the \hkl{100}, the \hkl{110} and the \hkl{111} surface. Red atom
           is the migrating atom, and its final position is the empty site 
           one nearest neighbor distance to the \hkl[110] direction (to the right in the last three panels).}
  \label{fig:surfaces}
\end{figure}

Splitting of the data in three subsets arises from the method in which
the barriers were calculated. See Figure~\ref{fig:surfaces} for an illustration
of the surface orientations in Cu crystal. For the calculation
of the migration energy barriers, each 26-dimensional input vector
was embedded in one of the three lowest index surfaces, according to a selected
criterion of stability. This results in different relaxation and forces being
present in the system, depending on the selected surface orientation. Furthermore,
in an earlier study using multilayer perceptrons for the same data set, accuracy
was gained by taking advantage of this physical split in the data~\cite{jyri1}.

We use the bracketed three-number Miller indexing for noting crystal orientations~\cite{kittel1996introduction}.
Briefly, the three numbers $h$, $k$ and $l$ are vector coordinates in the basis of cubic lattice vectors $\mathbf{a}_i$:
\hkl[hkl] is shorthand for $h\cdot \mathbf{a}_1 + k\cdot\mathbf{a}_2 + l\cdot\mathbf{a}_3$.
Overline signifies a negative number. Square brackets are used for vectors, angular brackets
for sets of equivalent vectors, round brackets for surfaces, and curly brackets
for sets of equivalent surfaces. The surface \hkl(hkl) can be defined as perpendicular
to the vector \hkl[hkl].

We compare the performance of our gradient-based RBF networks to two baselines:
gradient-boosted decision trees~\cite{gb}, as implemented in
the popular XGBoost~\cite{xgboost} package,
and deep neural networks implemented with the PyTorch framework.
For both baselines,
we use early stopping
after ten successive iterations of no improvement, and for XGBoost we impose an
upper limit of $10\,000$~trees. For each surface and both baseline algorithms, we first do 100~rounds
of hyperparameter tuning with random search~\cite{randomhp}.

For XGBoost, we
sample the learning rate uniformly from $[0.01,0.09]$, the instance
and column subsampling ratios from $\{0.5,0.6,\ldots,1.0\}$, the
maximum tree depth from $\{3,4,\ldots,15\}$, the minimum number of
instances in a node from $\{1,2,\ldots,10\}$, and the $L_1$ and $L_2$ regularization
weights and the minimum loss reduction required for a new partition
from $\{0,0.01,0.1,1,10\}$.

For deep neural networks, we sample
the learning rate from $\{ 10^{-2}, 10^{-3}, 10^{-4} \}$,
the batch size from $\{ 32, 64, 128, 256 \}$,
the $L_2$ regularization weight from $\{ 0 \} \cup \{  10^{-k} \colon k=1,2,\ldots,6  \}$,
the number of hidden layers from $\{ 3, 4, \ldots, 10\}$,
and
the number of nodes per hidden layer from $\{ 32, 64, 128, 256 \}$.
For each network, we randomly choose the activation function to be
either the widely used ReLU or the more recently proposed ELU~\cite{elu}.

For each sampled set of hyperparameters,
we perform ten random train--validation--test splits and use the
average of the test set MSE to rank the hyperparameter values.

\begin{figure}%
\centering%
\includegraphics[width=0.33\textwidth]{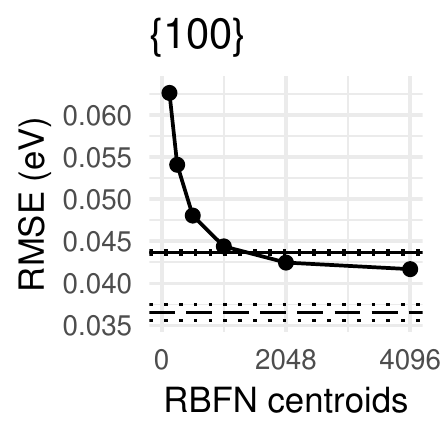}%
\includegraphics[width=0.33\textwidth]{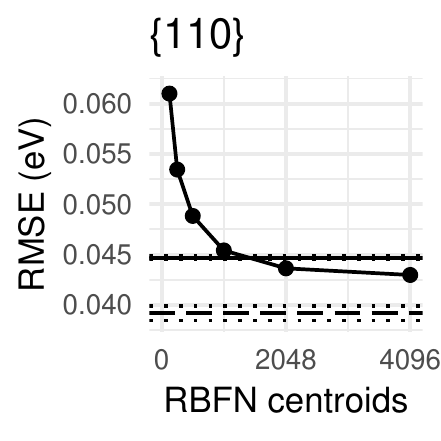}%
\includegraphics[width=0.33\textwidth]{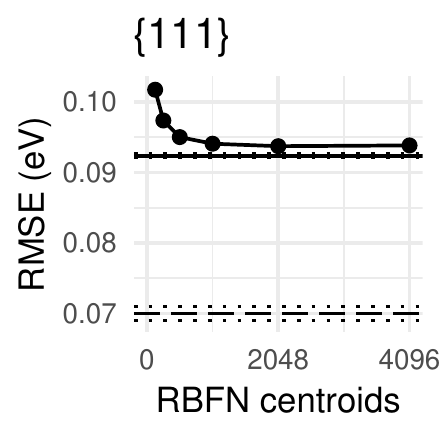}%
\caption{%
	Predictive performance of RBF networks (as a function of the number of centroids),
	XGBoost (shown by the solid horizontal line), and
	DNN (shown by the dashed line)
	on the three surfaces
	of the migration energy barrier data set.
	Error bars mostly not visible for RBFN's; dotted lines show one standard deviation for XGBoost and DNN.
	Note the different vertical axes.}%
\label{fig:xgb_rbfn}%
\end{figure}

We then evaluate the performance of the hyperparameter-tuned XGBoost and DNN
models and RBF networks with 128, 256, 512, 1024, 2048, and 4096 centroids,
each with ten random train--validation--test splits for each three
surfaces. The resulting root mean squared errors~(RMSE) are shown
in Table~\ref{table:rmse}
and also shown in Figure~\ref{fig:xgb_rbfn}.
(We also tried 8192-centroid RBF networks, but the improvements were insignificant.)
Overall, deep neural networks outperform the other models;
for all three surfaces, the best-performing DNN's use 256~nodes per layer, but the other optimized hyperparameters
found vary by surface (with, e.g., the number of layers ranging from 6~to~9).
For the \hkl{100} and \hkl{110}~surfaces,
the 4096-centroid RBFN is better than XGBoost, and the predictive
performance is acceptable also for the \hkl{111}~surface. The best
XGBoost predictors used 7324, 3419, and 1993 decision trees for the
\hkl{100}, \hkl{110}, and \hkl{111} surfaces, respectively.
The best hyperparameter values found for XGBoost lie either in the interiors of the predefined ranges or
at zero, which suggests that the hyperparameter search had sufficient coverage.

\begin{table}
\centering{}%
\begin{tabular}{ccc}
\toprule 
Surface & Model & RMSE (\textpm{} one s.d.)\tabularnewline
\midrule
\midrule 
\multirow{7}{*}{\hkl{100}} & RBFN~(128) & 0.0626 \textpm{} 0.0006\tabularnewline
\cmidrule{2-3} 
 & RBFN~(256) & 0.0541 \textpm{} 0.0004\tabularnewline
\cmidrule{2-3} 
 & RBFN~(512) & 0.0481 \textpm{} 0.0004\tabularnewline
\cmidrule{2-3} 
 & RBFN~(1024) & 0.0444 \textpm{} 0.0003\tabularnewline
\cmidrule{2-3} 
 & RBFN~(2048) & 0.0425 \textpm{} 0.0002\tabularnewline
\cmidrule{2-3} 
 & RBFN~(4096) & 0.0417 \textpm{} 0.0001\tabularnewline
\cmidrule{2-3} 
 & XGBoost & 0.0437 \textpm{} 0.0002\tabularnewline
\cmidrule{2-3} 
 & DNN & \textbf{0.0366 \textpm{} 0.0009}\tabularnewline
\midrule 
\multirow{7}{*}{\hkl{110}} & RBFN~(128) & 0.0610 \textpm{} 0.0006\tabularnewline
\cmidrule{2-3} 
 & RBFN~(256) & 0.0534 \textpm{} 0.0004\tabularnewline
\cmidrule{2-3} 
 & RBFN~(512) & 0.0488 \textpm{} 0.0004\tabularnewline
\cmidrule{2-3} 
 & RBFN~(1024) & 0.0454 \textpm{} 0.0003\tabularnewline
\cmidrule{2-3} 
 & RBFN~(2048) & 0.0436 \textpm{} 0.0002\tabularnewline
\cmidrule{2-3} 
 & RBFN~(4096) & 0.0430 \textpm{} 0.0001\tabularnewline
\cmidrule{2-3} 
 & XGBoost & 0.0447 \textpm{} 0.0002\tabularnewline
\cmidrule{2-3} 
 & DNN & \textbf{0.0392 \textpm{} 0.0007}\tabularnewline
\midrule 
\multirow{7}{*}{\hkl{111}} & RBFN~(128) & 0.1017 \textpm{} 0.0002\tabularnewline
\cmidrule{2-3} 
 & RBFN~(256) & 0.0974 \textpm{} 0.0003\tabularnewline
\cmidrule{2-3} 
 & RBFN~(512) & 0.0950 \textpm{} 0.0004\tabularnewline
\cmidrule{2-3} 
 & RBFN~(1024) & 0.0941 \textpm{} 0.0003\tabularnewline
\cmidrule{2-3} 
 & RBFN~(2048) & 0.0937 \textpm{} 0.0002\tabularnewline
\cmidrule{2-3} 
 & RBFN~(4096) & 0.0938 \textpm{} 0.0002\tabularnewline
\cmidrule{2-3} 
 & XGBoost & 0.0924 \textpm{} 0.0002\tabularnewline
\cmidrule{2-3} 
 & DNN & \textbf{0.0700 \textpm{} 0.0010}\tabularnewline
\cmidrule{2-3} 
\end{tabular}\caption{Predictive performance of RBF networks, XGBoost, and deep neural networks on the three surfaces
(sub--data sets) of the migration energy barrier data set, as measured
by root mean squared error (RMSE). The best model for each surface
is shown in boldface.}
\label{table:rmse}
\end{table}

\subsubsection{Visualization and interpretability by pruning}

To demonstrate the usefulness of our pruning approach, we first take
the best-performing RBF networks from Section~\ref{subsec:rbfn_vs_xgb}
for each surface and prune them down to sixteen centroids. Note that this
is different from directly training a sixteen-centroid RBFN: our goal
is to produce an interpretable approximation of the large model, and
this approximation is not meant to be used for producing predictions.
The predictive accuracy of a directly trained sixteen-centroid model
would be better than that of the pruned model
but much worse than that of a large RBFN with thousands of centroids;
its centroids would be optimized for prediction,
not for approximating a larger RBF network and aiding in its interpretation.

In this case, we use the Bernoulli distribution with $p=0.5$ for
each of the 26~features. Hence, denoting the large RBFN by $f$ and
the pruned one by $f'$, we are essentially solving the optimization
problem
\[
\arg\min_{f'}\left\{ \frac{1}{2^{26}}\sum_{\bm{x}\in\{-1,1\}^{26}}\left[f(\bm{x})-f'(\bm{x})\right]^{2}\right\} \text{,}
\]
which seemingly involves $2^{26}$~terms but can be optimized efficiently
because of Theorem~\ref{thm:pruning_expectations}. As before, we
use ten random initializations for the pruning and select the RBFN
that gives the best result.
The square roots of the resulting pruning
loss values~\eqref{eq:e_decomposition} are
0.0650, 0.0653, and 0.0908 for the surfaces \hkl{100}, \hkl{110}, and \hkl{111}, respectively.
While these values naturally show that the heavy pruning incurs a reduction in accuracy,
the values are nevertheless reasonably good when one considers that the response values
in the data sets have the ranges
$0.41 \pm 0.37$,
$0.43 \pm 0.31$, and
$0.43 \pm 0.32$
($\pm$~shows one standard deviation).

Interestingly, we can visualize the centroids of the resulting RBF
networks and gain insight into the structure of the prediction task
and the workings of the RBF network.
All visualizations for the centroids of the pruned RBFN's are shown in Figures~\ref{fig:centroids100a}--\ref{fig:centroids111b}
in \ref{sec:centroids}.
The 26~elements of~$\Theta_i$ corresponding to each centroid~$i$ in the pruned RBFN are shown by the color and the opacity of the atomic positions:
red encodes a positive value, blue a negative value, and the opacity is proportional to the absolute value of the element
(fully opaque means a \mbox{value $\geq 2$}). The same prototype is shown from four different viewpoints: \hkl[1-10],
\hkl[001], \hkl[-1-10], and \hkl[0-31].

\begin{figure}
  \centering
  \begin{subfigure}{0.24\linewidth}
    \centering
    \includegraphics[width=\linewidth]{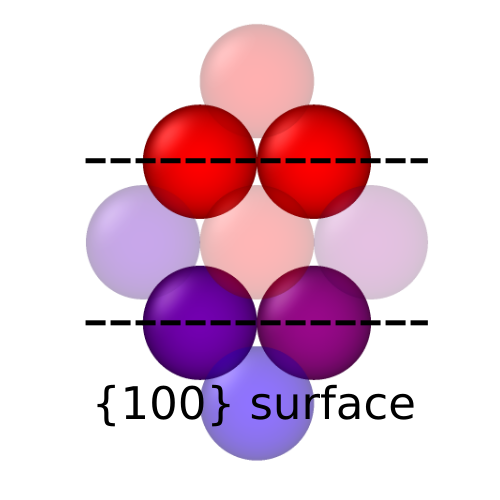}
    \caption{\hkl{100} (10)}
  \end{subfigure}
  \begin{subfigure}{0.24\linewidth}
    \centering
    \includegraphics[width=\linewidth]{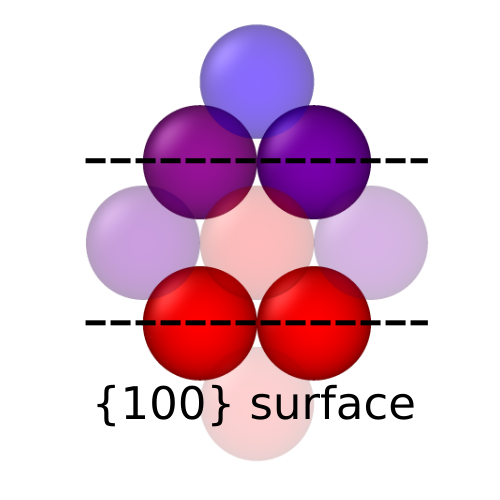}
    \caption{\hkl{100} (14)}
  \end{subfigure}
  \begin{subfigure}{0.24\linewidth}
    \centering
    \includegraphics[width=\linewidth]{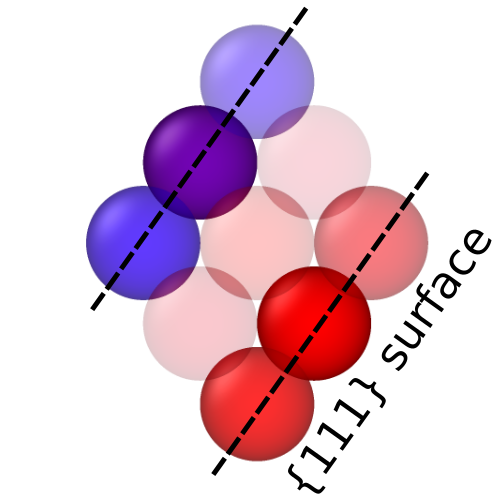}
    \caption{\hkl{111} (1)}
  \end{subfigure}
  \begin{subfigure}{0.24\linewidth}
    \centering
    \includegraphics[width=\linewidth]{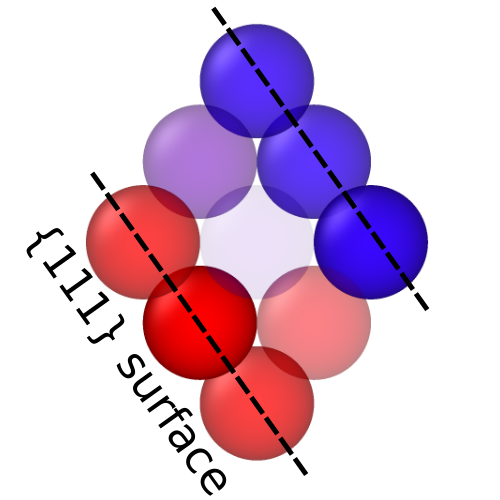}
    \caption{\hkl{111} (6)}
  \end{subfigure}
  \caption{Prototypes from the \hkl{100} and the \hkl{111} surfaces that display high opacity patterns parallel to their respective surface orientations.}
  \label{fig:symmetries}
\end{figure}

Several patterns related to the surfaces' orientations can be observed from the distributions of opacities. Namely, in the \hkl{100} prototypes, more than in the other groups, a pattern of high opacity aligning with the \hkl{100} surface can be seen. Likewise, many of the \hkl{111} set
prototypes have structures aligning with the \hkl{111} surface. Illustrative examples of prototypes from the \hkl{100} and the \hkl{111} networks are depicted in Figure~\ref{fig:symmetries}.

\begin{figure}
  \centering
  \begin{subfigure}{0.32\linewidth}
    \centering
    \includegraphics[width=\linewidth]{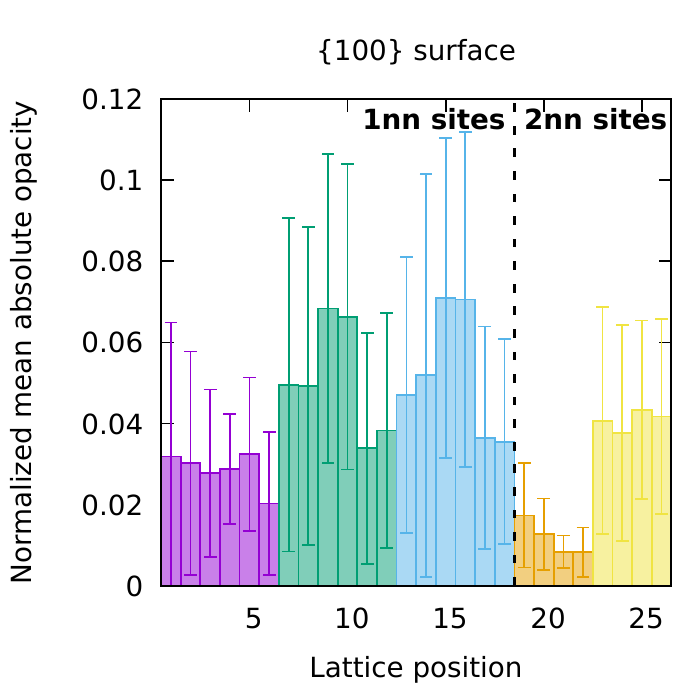}
  \end{subfigure}
  \begin{subfigure}{0.32\linewidth}
    \centering
    \includegraphics[width=\linewidth]{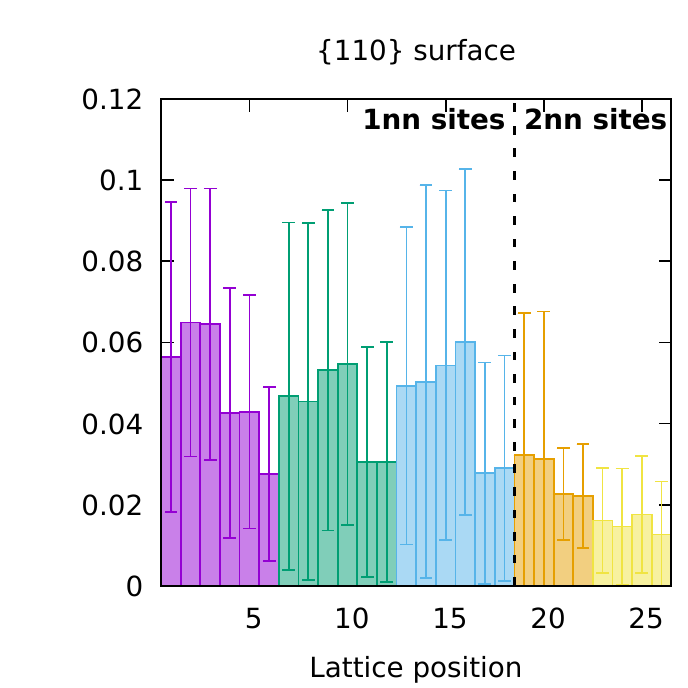}
  \end{subfigure}
  \begin{subfigure}{0.32\linewidth}
    \centering
    \includegraphics[width=\linewidth]{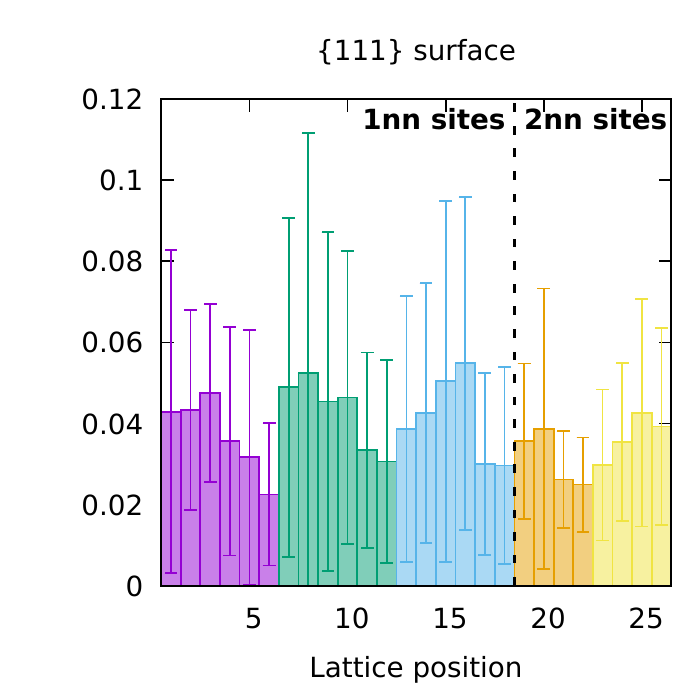}
  \end{subfigure}
  \caption{Normalized mean absolute opacity of different lattice position in each pruned RBF network.
           Error bars are standard deviations. Coloring is according to groups of lattice positions in
           different layers of the atomic environment; see Figure~\ref{fig:surfaces} for illustration.}
  \label{fig:histograms}
\end{figure}

Relevant physical features contained in the RBFN's can be also inspected from the mean absolute opacities
of each lattice site. This information is plotted in Figure~\ref{fig:histograms}.
Color coding is the same as in Figure~\ref{fig:surfaces}: by the atomic layer and the distance to the migrating atom. 

The first clear observation is that the first nearest neighbor (1nn) sites on average have higher opacities as
the pruned RBFN's are small and incorporate only the most important patterns. Another notable feature is
that \hkl{110} prototypes have lower opacities on average on the last four lattice positions. This, again,
corresponds to the lattice structure, as on the \hkl{110} surface, these four positions are located in the same
layer as the migrating atom, but quite far away, at second nearest neighbor (2nn) distance, on the neighboring ridges.
See the second-rightmost panel of Figure~\ref{fig:surfaces} for an illustration: the last four sites are the light
yellow positions at the top and the bottom of that figure.
Sites 19--22 on the \hkl{100} surface have a similar role, and these sites indeed have the lowest mean opacity.
It should be expected that 2nn sites in the same layer with the migrating atom have the least contribution to the migration barrier,
since they are far from the migrating atom, and their absence or presence does not even impose much stress
on the system, unlike that of the sites in the lower layers. On the \hkl{111} surface, there are no 2nn sites in the same layer with the migrating
atom---this is reflected in the more even opacity of all lattice site groups.

The distributions of lattice position opacities on different surfaces suggest that the RBFN's were able to infer physically
meaningful information only from the functional dependency between binary input and the migration energy barriers, without ever having
access to three-dimensional representations of the lattice sites.

The input patterns that the RBFN algorithm found to be meaningful for the regression model may prove useful
in developing more sophisticated input descriptors for future machine learning solutions of similar problems.
The currently used integer vector descriptor, developed by Djurabekova et al.~\cite{djurabekova2007stability,
djurabekova2007artificial}, has certain limitations: while it is invariant in three-dimensional
translations of the lattice (arising from the total omission of the three-dimensional coordinates),
it is \emph{not} invariant in reflection about the \hkl(001) or the \hkl(1-10) planes (see the second panel
of Figure~\ref{fig:surfaces} for the vectors normal to these planes) and 180\,\degree\ rotation around the
\hkl[110] axis. The migration energy barrier response $E_\mathrm{m}$ used in KMC simulations has to
be invariant in these reflections and rotations to give similar diffusion properties in physically
equivalent local atomic environments on differently oriented surfaces.

The lack of important invariances in the descriptor can be circumvented e.g. by systematically
choosing a representative input from each family of physically equivalent cases, as was done in ref.~\cite{jyri2},
or by averaging over all the symmetric cases when producing the response for an input.
The former method will waste some of the available training data, and prevent the regressor
from learning the symmetries itself, while the latter method will spend nearly four times
as much resources at each function call. A properly invariant descriptor would render these
workarounds unnecessary and could potentially even reduce the dimensionality of the input space.

We are aware of some descriptors popularly used in representing atomistic input data, such
as the smooth overlap of atomic positions (SOAP) descriptor~\cite{bartok2013representing}, that are invariant in
translations, rotations, and reflections. These descriptors have been
developed for a somewhat different task than ours---for mapping atomistic
input to \emph{total energy} of the system, and often also the forces present in it,
as opposed to just a single migration energy barrier of a transition process.
The different problem setting motivates different properties for the descriptor.
The SOAP descriptor in particular was developed to give a continuous similarity kernel
for comparing systems where atomic positions are not restricted to certain lattice positions.
The migration energy problem can certainly be formulated in terms of total energies:
$E_\mathrm{m}$ is the difference between the energy of the system at its saddle position
(roughly, halfway through the jump) and its initial position. For the purpose of KMC simulations
only, the total energies can be expensive extra information, as the only parameters of interest
are the barriers. Nevertheless, at least Messina et al. have
taken this approach using the bispectrum descriptor~\cite{messina2018smart}.

While the SOAP descriptor specifically may be unnecessarily complicated for a
rigid-lattice system, where its continuity properties are not needed, the applications
of these kinds of descriptors in the direct migration barrier regression could be
explored in future work. At the same time, developing new descriptors precisely
suited for this task might prove fruitful.
Relevant input patterns, such as those discovered in this work, can guide in the design
of descriptors that have the desired invariance properties. One could imagine a set of templates,
created either manually or as a part of the training process, that are convoluted over the
three-dimensional representation of the local atomic environment, to produce the input values
fed to a machine learning regressor.

\section{Conclusions}

\label{sec:end}

The radial basis function network (RBFN) is a classic model for supervised machine learning
and has good interpretability properties when the number of centroids is small.
For training RBFN's, previous research has mainly concentrated on heuristic two-step methods,
apparently because gradient descent--based optimization for small RBFN's has faced local minima problems.

In this article, we introduced a new PyTorch-based RBFN implementation
that can be combined with modern optimization techniques
that are currently used in deep learning.
With a large number of centroids and full gradient-based optimization,
we showed that RBFN's can achieve predictive performance on~par with
hyperparameter-optimized gradient-boosted decision trees
on a materials physics data set that describes migration energy barriers for atom configurations on three different copper surfaces.

To make the trained RBF networks interpretable, we derived a novel pruning method
based on finding a smaller RBFN that globally approximates the larger one
given a suitable assumption on the distribution of the data manifold.
We provided closed-form pruning objective functions for the cases where the input features
are assumed to follow either a Bernoulli, a continuous uniform, or a Gaussian mixture distribution.
Using the Bernoulli objective,
we pruned RBFN's with thousands of centroids trained on the aforementioned materials physics data set
down to sixteen centroids. Visualizations of the obtained centroids
show that the large RBFN's have learned multiple patterns that match the properties of the physical lattice
without the models having had access to any \emph{a~priori} knowledge
on the nature of the problem.

While our methods produce good results on a real-world data set, many possibilities remain for further refinement.
We have only considered the common but quite restrictive exponential kernel, and
one can also generalize the functional form of RBF networks in various ways.
For instance, having a separate~$\gamma_i$ parameter for each centroid would increase the flexibility of the model,
though at the expense of making at least the pruning objective derivations more complicated.
For pruning, one could consider automatically approximating the distribution of the training data set
with Gaussian mixture--based kernel density estimators, though this would entail the risk of overfitting and
hence producing pruned RBFN's that are unable to generalize from the training data.

\section*{Acknowledgments}

This work was supported by the Academy of Finland (projects 313857
and 313867). Computational resources were provided by the Finnish
Grid and Cloud Infrastructure \mbox{(urn:nbn:fi:research-infras-2016072533)}.

\bibliographystyle{elsarticle-num}
\bibliography{main}

\appendix

\section{Proof of Theorem~\ref{thm:pruning_expectations}}
\label{sec:proof}

\begin{proof}
Let $k,r\geq0$ and $\bm{u},\bm{v}\in\mathbb{R}^{D}$. Assume first
that $\bm{x}$ has the mixture density
\[
p(\bm{x})=\prod_{i=1}^{D}\sum_{j=1}^{L_{i}}w_{ij}\:f_{N}(x_{i}\mid\mu_{ij},\sigma_{ij}^{2})\text{.}
\]
Then
\[
\mathbb{E}_{p(\bm{x})}\left[e^{-k\|\bm{x}-\bm{u}\|^{2}-r\|\bm{x}-\bm{v}\|^{2}}\right]=\prod_{i=1}^{D}\sum_{j=1}^{L_{i}}w_{ij}\,\mathbb{E}_{x_{i}\sim N(\mu_{ij},\sigma_{ij}^{2})}\left[e^{-k(x_{i}-u_{i})^{2}-r(x_{i}-v_{i})^{2}}\right]\text{,}
\]
and by straightforward algebraic manipulation, we find that the right-hand-side
expectation equals
\begin{align*}
 & \mathbb{E}_{x_{i}\sim N(\mu_{ij},\sigma_{ij}^{2})}\left[e^{-\gamma(x_{i}-y_{i})^{2}-\rho(x_{i}-z_{i})^{2}}\right]\\
= & \int_{\mathbb{R}}(2\pi\sigma_{ij}^{2})^{-1/2}\,e^{-\frac{(x_{i}-\mu_{ij})^{2}}{2\sigma_{ij}^{2}}}\:e^{-k(x_{i}-u_{i})^{2}-r(x_{i}-v_{i})^{2}}\,\textrm{d}x_{i}\\
= & \left(1+2\sigma_{ij}^{2}(k+r)\right)^{-1/2}\exp\left(-\frac{k(u_{i}-\mu_{ij})^{2}+r(v_{i}-\mu_{ij})^{2}+2\sigma_{ij}^{2}kr(u_{i}-v_{i})^{2}}{1+2\sigma_{ij}^{2}(k+r)}\right)\\
 & \quad\times\int_{\mathbb{R}}\left(\frac{2\pi\sigma_{ij}^{2}}{1+2\sigma_{ij}^{2}(k+r)}\right)^{-1/2}\,\exp\left(-\frac{\left(x-\frac{\mu_{ij}+2\sigma_{ij}^{2}(ku_{i}+rv_{i})}{1+2\sigma_{ij}^{2}(k+r)}\right)^{2}}{2\left(\frac{\sigma_{ij}^{2}}{1+2\sigma_{ij}^{2}(k+r)}\right)}\right)\,\textrm{d}x_{i}\text{.}
\end{align*}
By recognizing that the final integral equals one, we obtain the desired
form.

Consider then the uniform density
\[
p(\bm{x})=\prod_{i=1}^{D}f_{U}(x_{i}\mid a_{i},b_{i})\text{.}
\]
As previously, we decompose
\[
\mathbb{E}_{p(\bm{x})}\left[e^{-k\|\bm{x}-\bm{u}\|^{2}-r\|\bm{x}-\bm{v}\|^{2}}\right]=\prod_{i=1}^{D}\mathbb{E}_{x_{i}\sim U(a_{i},b_{i})}\left[e^{-k(x_{i}-u_{i})^{2}-r(x_{i}-v_{i})^{2}}\right]\text{,}
\]
and by algebraic manipulation we arrive at a form involving the Gaussian
CDF which we express in terms of the error function:
\begin{align*}
 & \mathbb{E}_{x_{i}\sim U(a_{i},b_{i})}\left[e^{-k(x_{i}-u_{i})^{2}-r(x_{i}-v_{i})^{2}}\right]\\
= & \frac{1}{b_{i}-a_{i}}\int_{a_{i}}^{b_{i}}e^{-k(x_{i}-u_{i})^{2}-r(x_{i}-v_{i})^{2}}\,\textrm{d}x_{i}\\
= & \frac{1}{b_{i}-a_{i}}\,\sqrt{\frac{\pi}{k+r}}\,\exp\left(-\frac{kr(u_{i}-v_{i})^{2}}{k+r}\right)\\
 & \quad\times\int_{a_{i}}^{b_{i}}\left(\frac{2\pi}{2(k+r)}\right)^{-1/2}\exp\left(-\frac{\left(x_{i}-\frac{ku_{i}+rv_{i}}{k+r}\right)^{2}}{2\left(\frac{1}{2(k+r)}\right)}\right)\,\textrm{d}x_{i}\\
= & \frac{1}{b_{i}-a_{i}}\,\sqrt{\frac{\pi}{k+r}}\,\exp\left(-\frac{kr(u_{i}-v_{i})^{2}}{k+r}\right)\\
 & \quad\times\frac{1}{2}\left[\textrm{erf}\left(\frac{k(b_{i}-u_{i})+r(b_{i}-v_{i})}{\sqrt{k+r}}\right)-\textrm{erf}\left(\frac{k(a_{i}-u_{i})+r(a_{i}-v_{i})}{\sqrt{k+r}}\right)\right]\text{.}
\end{align*}
This concludes the uniform case.

Finally, consider the binary case
\[
p(\bm{x})=\prod_{i=1}^{D}p_{\textrm{Bernoulli}}(x_{i}\mid q)\text{,}
\]
where $p_{\textrm{Bernoulli}}(\cdot\mid q)$ is the Bernoulli point
probability function for the outcomes~$\pm1$. We denote the first
$n$~elements of~$\bm{x}$ by~$\bm{x}_{1:n}$ (and similarly for
other vectors) and give a proof by induction on the number of dimensions.
The case $D=1$ is straightforward:
\begin{align*}
 & \mathbb{E}_{p(\bm{x}_{1:1})}\left[e^{-k\|\bm{x}_{1:1}-\bm{u}_{1:1}\|^{2}-r\|\bm{x}_{1:1}-\bm{v}_{1:1}\|^{2}}\right]\\
=\, & \mathbb{E}_{p_{\textrm{Bernoulli}}(x_{1}\mid q_{1})}\left[e^{-k(x_{1}-u_{1})^{2}-r(x_{1}-v_{1})^{2}}\right]\\
=\, & q_{1}\,e^{-k(1-u_{1})^{2}-r(1-v_{1})^{2}}+(1-q_{1})\,e^{-k(-1-u_{1})^{2}-r(-1-v_{1})^{2}}\\
=\, & e^{-k(u_{1}+1)^{2}-r(v_{1}+1)^{2}}\left[1+q_{1}\left(e^{4(ku_{1}+rv_{1})}-1\right)\right]\text{.}
\end{align*}
Assume then that the claim holds for some $D\in\mathbb{N}$ and consider
the case with $D+1$ dimensions. We have
\begin{align*}
 & \mathbb{E}_{p(\bm{x}_{1:D+1})}\left[e^{-k\|\bm{x}_{1:D+1}-\bm{u}_{1:D+1}\|^{2}-r\|\bm{x}_{1:D+1}-\bm{v}_{1:D+1}\|^{2}}\right]\\
=\, & \mathbb{E}_{p_{\textrm{Bernoulli}}(x_{D+1}\mid q_{D+1})}\left[e^{-k(x_{D+1}-u_{D+1})^{2}-r(x_{D+1}-v_{D+1})^{2}}\right]\\
 & \quad\times\mathbb{E}_{p(\bm{x}_{1:D})}\left[e^{-k\|\bm{x}_{1:D}-\bm{u}_{1:D}\|^{2}-r\|\bm{x}_{1:D}-\bm{v}_{1:D}\|^{2}}\right]\\
=\, & e^{-k(u_{D+1}+1)^{2}-r(v_{D+1}+1)^{2}}\left[1+q_{D+1}\left(e^{4(ku_{D+1}+rv_{D+1})}-1\right)\right]\\
 & \quad\times e^{-k\|\bm{u}_{1:D}+\bm{1}\|^{2}-r\|\bm{v}_{1:D}+\bm{1}\|^{2}}\prod_{i=1}^{D}\left[1+q_{i}\left(e^{4\left(ku_{i}+rv_{i}\right)}-1\right)\right]\\
=\, & e^{-k\|\bm{u}_{1:D+1}+\bm{1}\|^{2}-r\|\bm{v}_{1:D+1}+\bm{1}\|^{2}}\prod_{i=1}^{D+1}\left[1+q_{i}\left(e^{4\left(ku_{i}+rv_{i}\right)}-1\right)\right]
\end{align*}
which completes the proof.
\end{proof}

\section{Centroids of the pruned RBFN's}
\label{sec:centroids}

\begin{figure}%
\raggedleft%
\raisebox{-.5\height}{\includegraphics[width=0.22\textwidth]{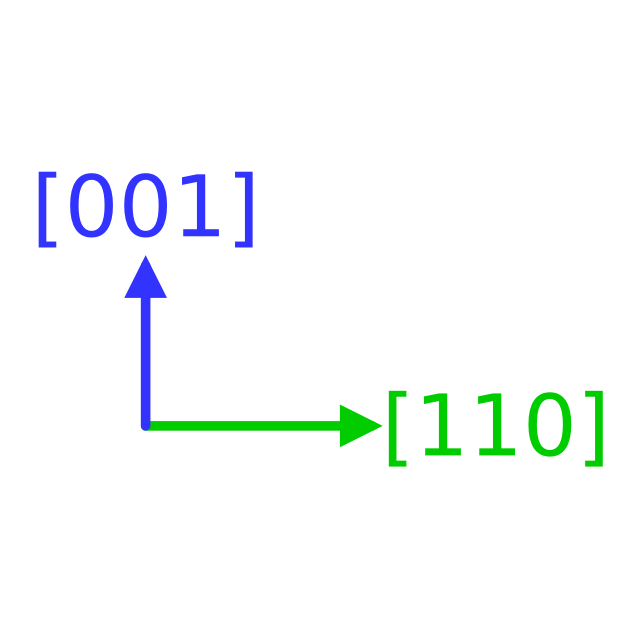}}%
\raisebox{-.5\height}{\includegraphics[width=0.22\textwidth]{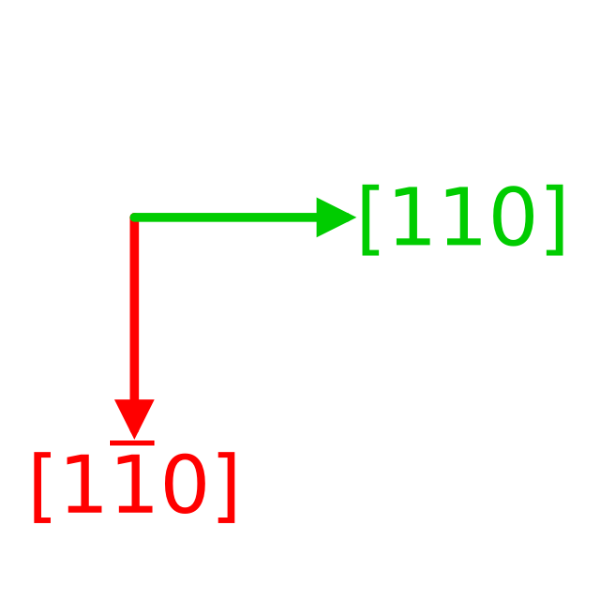}}%
\raisebox{-.5\height}{\includegraphics[width=0.22\textwidth]{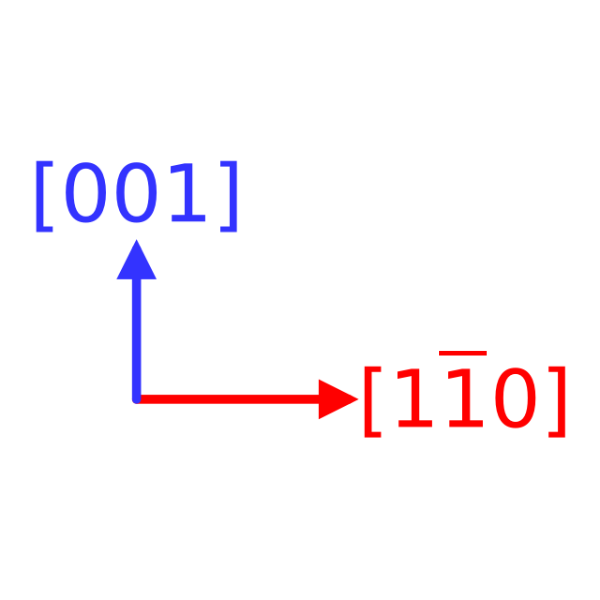}}%
\raisebox{-.5\height}{\includegraphics[width=0.22\textwidth]{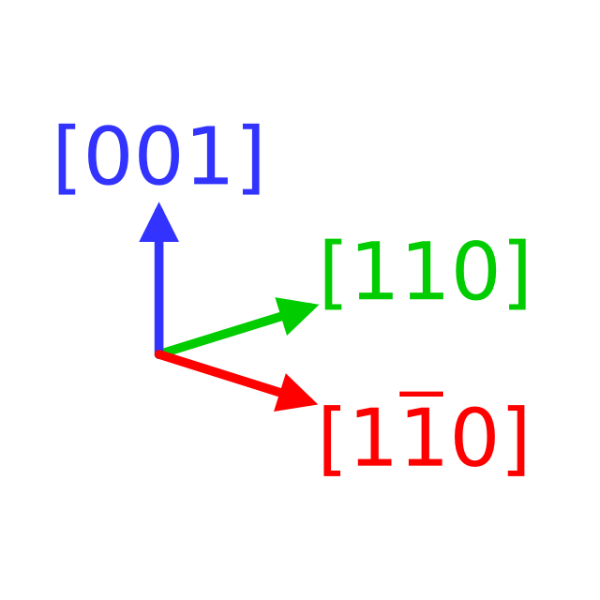}}%
\\%
(1)%
\raisebox{-.5\height}{\includegraphics[width=0.22\textwidth]{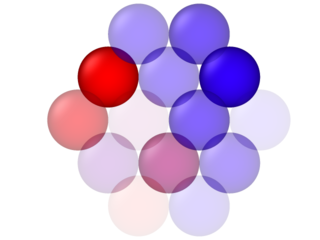}}%
\raisebox{-.5\height}{\includegraphics[width=0.22\textwidth]{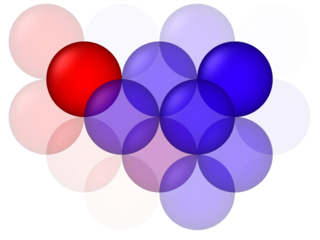}}%
\raisebox{-.5\height}{\includegraphics[width=0.22\textwidth]{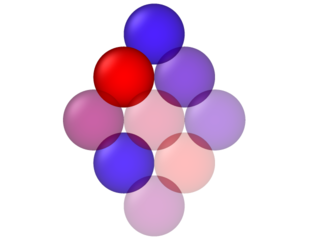}}%
\raisebox{-.5\height}{\includegraphics[width=0.22\textwidth]{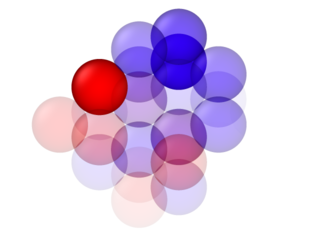}}%
\\%
(2)%
\raisebox{-.5\height}{\includegraphics[width=0.22\textwidth]{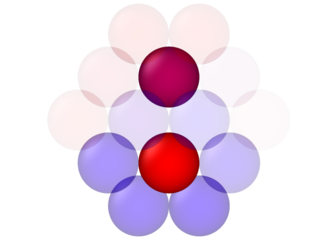}}%
\raisebox{-.5\height}{\includegraphics[width=0.22\textwidth]{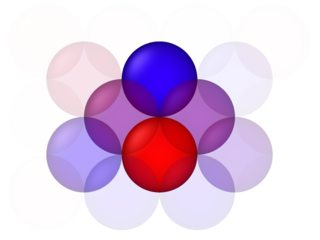}}%
\raisebox{-.5\height}{\includegraphics[width=0.22\textwidth]{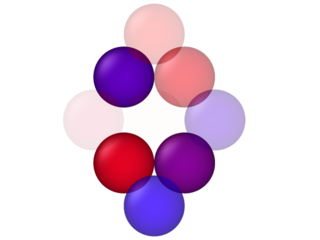}}%
\raisebox{-.5\height}{\includegraphics[width=0.22\textwidth]{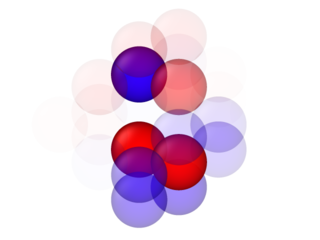}}%
\\%
(3)%
\raisebox{-.5\height}{\includegraphics[width=0.22\textwidth]{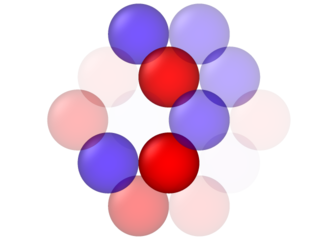}}%
\raisebox{-.5\height}{\includegraphics[width=0.22\textwidth]{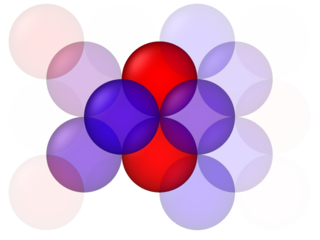}}%
\raisebox{-.5\height}{\includegraphics[width=0.22\textwidth]{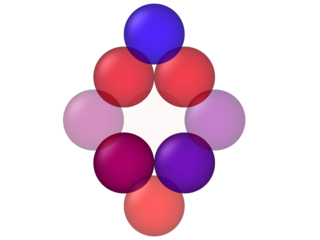}}%
\raisebox{-.5\height}{\includegraphics[width=0.22\textwidth]{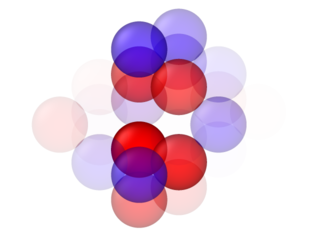}}%
\\%
(4)%
\raisebox{-.5\height}{\includegraphics[width=0.22\textwidth]{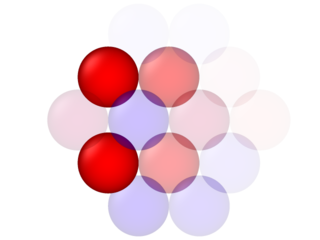}}%
\raisebox{-.5\height}{\includegraphics[width=0.22\textwidth]{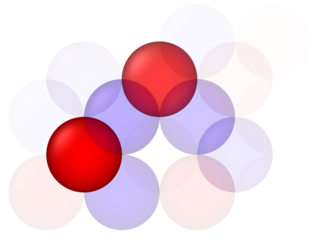}}%
\raisebox{-.5\height}{\includegraphics[width=0.22\textwidth]{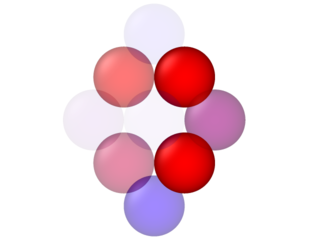}}%
\raisebox{-.5\height}{\includegraphics[width=0.22\textwidth]{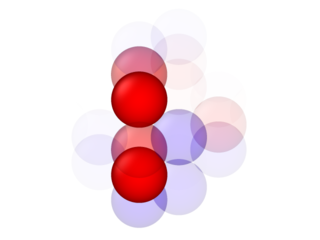}}%
\\%
(5)%
\raisebox{-.5\height}{\includegraphics[width=0.22\textwidth]{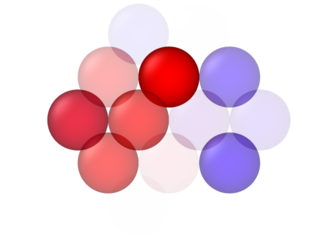}}%
\raisebox{-.5\height}{\includegraphics[width=0.22\textwidth]{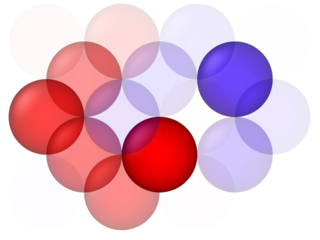}}%
\raisebox{-.5\height}{\includegraphics[width=0.22\textwidth]{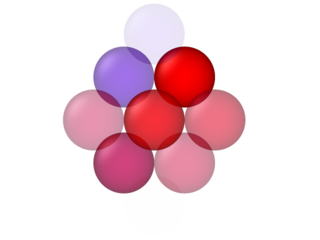}}%
\raisebox{-.5\height}{\includegraphics[width=0.22\textwidth]{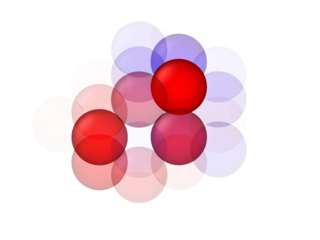}}%
\\%
(6)%
\raisebox{-.5\height}{\includegraphics[width=0.22\textwidth]{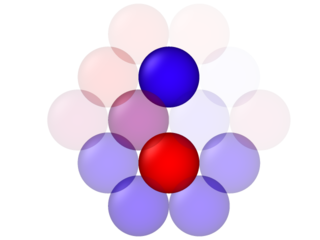}}%
\raisebox{-.5\height}{\includegraphics[width=0.22\textwidth]{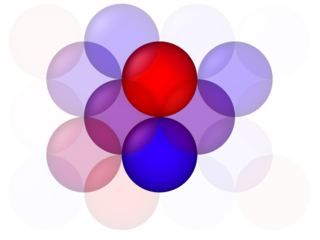}}%
\raisebox{-.5\height}{\includegraphics[width=0.22\textwidth]{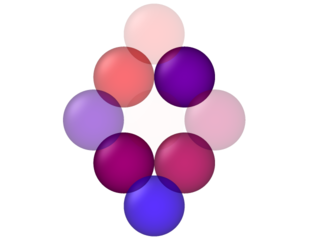}}%
\raisebox{-.5\height}{\includegraphics[width=0.22\textwidth]{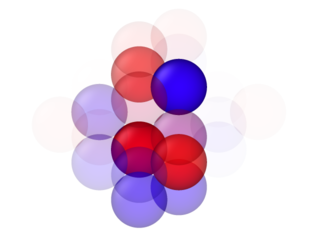}}%
\\%
(7)%
\raisebox{-.5\height}{\includegraphics[width=0.22\textwidth]{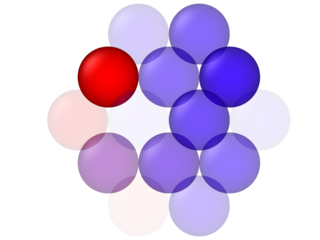}}%
\raisebox{-.5\height}{\includegraphics[width=0.22\textwidth]{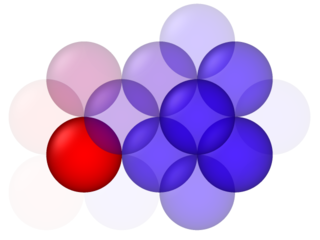}}%
\raisebox{-.5\height}{\includegraphics[width=0.22\textwidth]{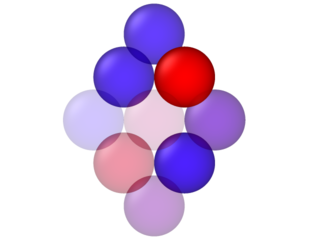}}%
\raisebox{-.5\height}{\includegraphics[width=0.22\textwidth]{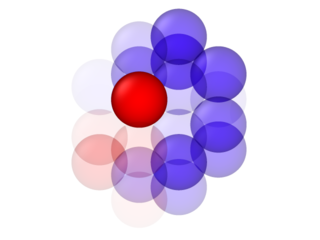}}%
\\%
(8)%
\raisebox{-.5\height}{\includegraphics[width=0.22\textwidth]{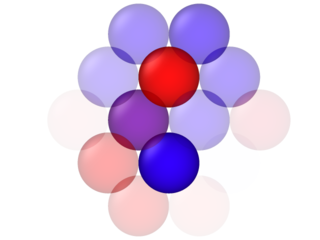}}%
\raisebox{-.5\height}{\includegraphics[width=0.22\textwidth]{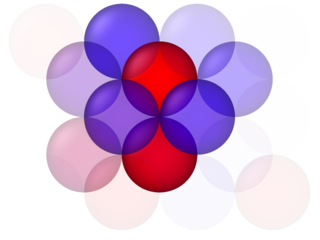}}%
\raisebox{-.5\height}{\includegraphics[width=0.22\textwidth]{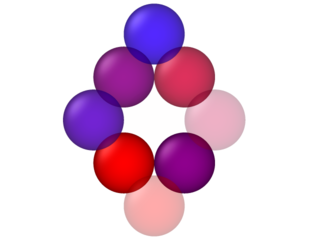}}%
\raisebox{-.5\height}{\includegraphics[width=0.22\textwidth]{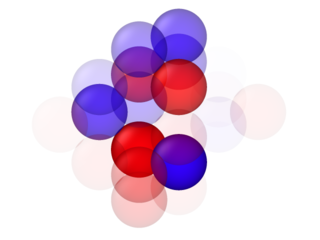}}%
\caption{Prototypes 1--8 of the \hkl{100} RBFN.}%
\label{fig:centroids100a}%
\end{figure}

\begin{figure}%
\raggedleft%
\raisebox{-.5\height}{\includegraphics[width=0.22\textwidth]{side_coordinates.png}}%
\raisebox{-.5\height}{\includegraphics[width=0.22\textwidth]{top_coordinates.png}}%
\raisebox{-.5\height}{\includegraphics[width=0.22\textwidth]{back_coordinates.png}}%
\raisebox{-.5\height}{\includegraphics[width=0.22\textwidth]{ortho_coordinates.png}}%
\\%
(9)%
\raisebox{-.5\height}{\includegraphics[width=0.22\textwidth]{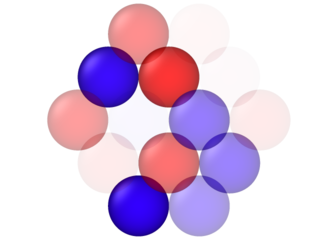}}%
\raisebox{-.5\height}{\includegraphics[width=0.22\textwidth]{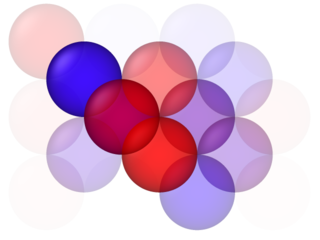}}%
\raisebox{-.5\height}{\includegraphics[width=0.22\textwidth]{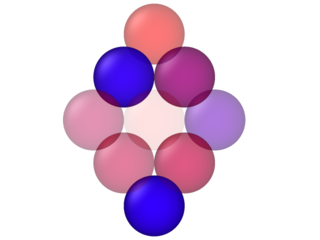}}%
\raisebox{-.5\height}{\includegraphics[width=0.22\textwidth]{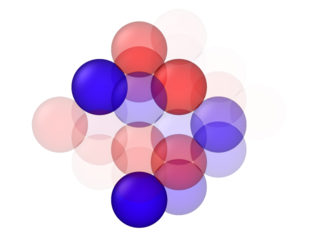}}%
\\%
(10)%
\raisebox{-.5\height}{\includegraphics[width=0.22\textwidth]{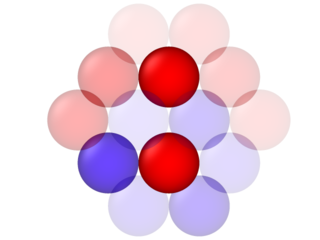}}%
\raisebox{-.5\height}{\includegraphics[width=0.22\textwidth]{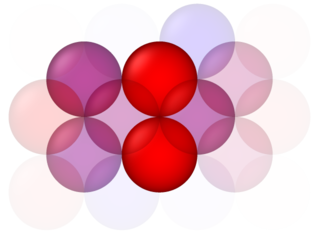}}%
\raisebox{-.5\height}{\includegraphics[width=0.22\textwidth]{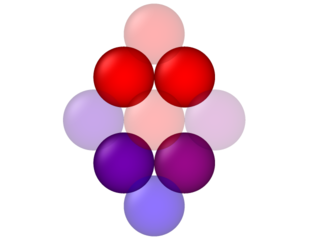}}%
\raisebox{-.5\height}{\includegraphics[width=0.22\textwidth]{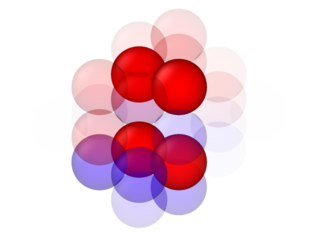}}%
\\%
(11)%
\raisebox{-.5\height}{\includegraphics[width=0.22\textwidth]{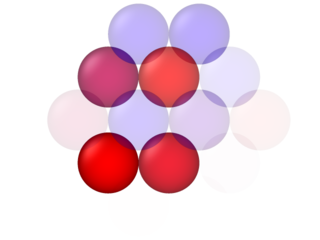}}%
\raisebox{-.5\height}{\includegraphics[width=0.22\textwidth]{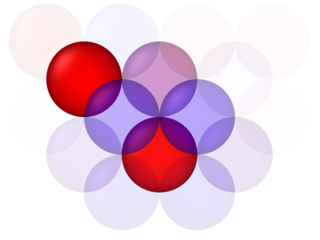}}%
\raisebox{-.5\height}{\includegraphics[width=0.22\textwidth]{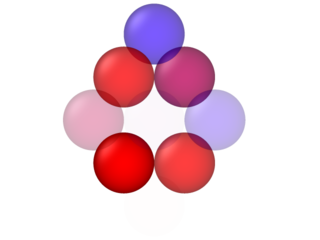}}%
\raisebox{-.5\height}{\includegraphics[width=0.22\textwidth]{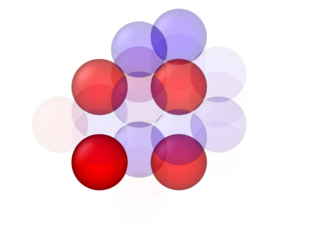}}%
\\%
(12)%
\raisebox{-.5\height}{\includegraphics[width=0.22\textwidth]{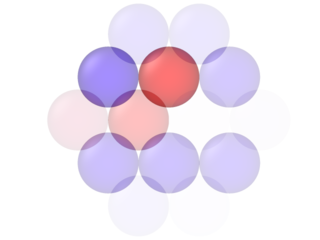}}%
\raisebox{-.5\height}{\includegraphics[width=0.22\textwidth]{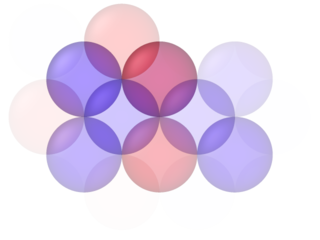}}%
\raisebox{-.5\height}{\includegraphics[width=0.22\textwidth]{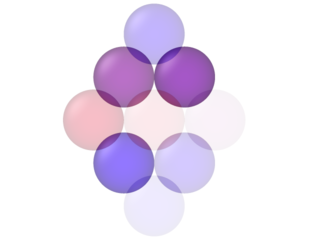}}%
\raisebox{-.5\height}{\includegraphics[width=0.22\textwidth]{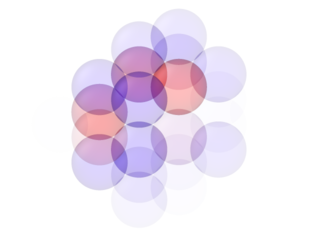}}%
\\%
(13)%
\raisebox{-.5\height}{\includegraphics[width=0.22\textwidth]{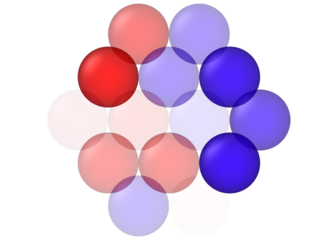}}%
\raisebox{-.5\height}{\includegraphics[width=0.22\textwidth]{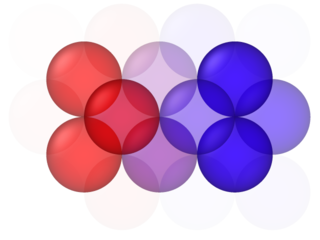}}%
\raisebox{-.5\height}{\includegraphics[width=0.22\textwidth]{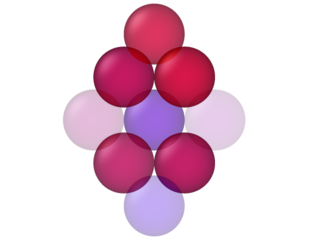}}%
\raisebox{-.5\height}{\includegraphics[width=0.22\textwidth]{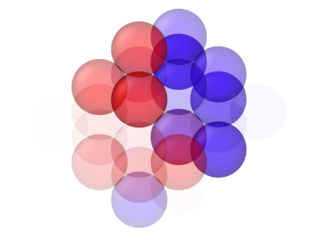}}%
\\%
(14)%
\raisebox{-.5\height}{\includegraphics[width=0.22\textwidth]{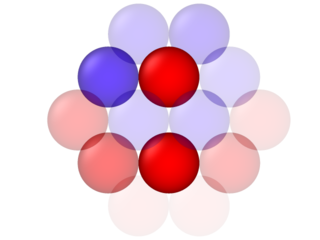}}%
\raisebox{-.5\height}{\includegraphics[width=0.22\textwidth]{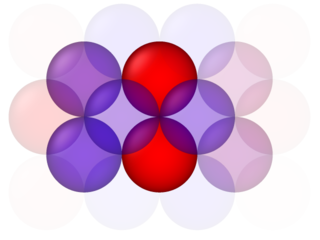}}%
\raisebox{-.5\height}{\includegraphics[width=0.22\textwidth]{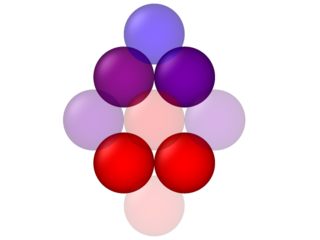}}%
\raisebox{-.5\height}{\includegraphics[width=0.22\textwidth]{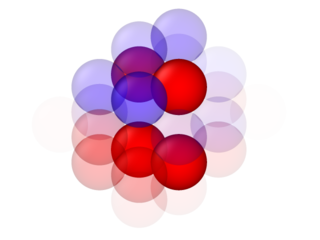}}%
\\%
(15)%
\raisebox{-.5\height}{\includegraphics[width=0.22\textwidth]{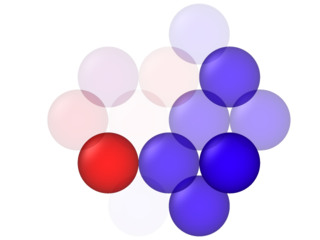}}%
\raisebox{-.5\height}{\includegraphics[width=0.22\textwidth]{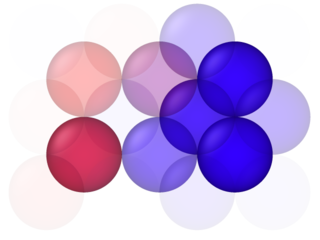}}%
\raisebox{-.5\height}{\includegraphics[width=0.22\textwidth]{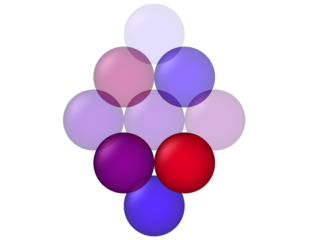}}%
\raisebox{-.5\height}{\includegraphics[width=0.22\textwidth]{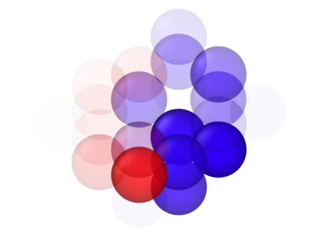}}%
\\%
(16)%
\raisebox{-.5\height}{\includegraphics[width=0.22\textwidth]{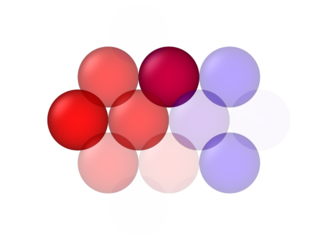}}%
\raisebox{-.5\height}{\includegraphics[width=0.22\textwidth]{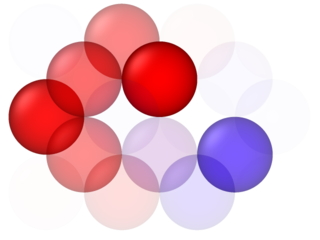}}%
\raisebox{-.5\height}{\includegraphics[width=0.22\textwidth]{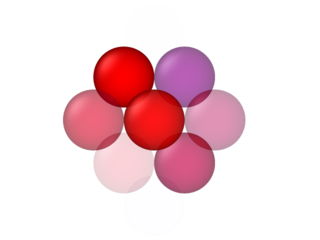}}%
\raisebox{-.5\height}{\includegraphics[width=0.22\textwidth]{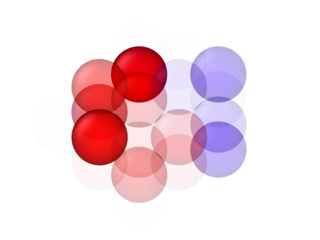}}%
\caption{Prototypes 9--16 of the \hkl{100} RBFN.}%
\label{fig:centroids100b}%
\end{figure}

\begin{figure}%
\raggedleft%
\raisebox{-.5\height}{\includegraphics[width=0.22\textwidth]{side_coordinates.png}}%
\raisebox{-.5\height}{\includegraphics[width=0.22\textwidth]{top_coordinates.png}}%
\raisebox{-.5\height}{\includegraphics[width=0.22\textwidth]{back_coordinates.png}}%
\raisebox{-.5\height}{\includegraphics[width=0.22\textwidth]{ortho_coordinates.png}}%
\\%
(1)%
\raisebox{-.5\height}{\includegraphics[width=0.22\textwidth]{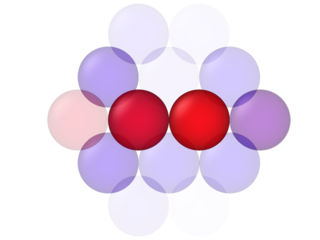}}%
\raisebox{-.5\height}{\includegraphics[width=0.22\textwidth]{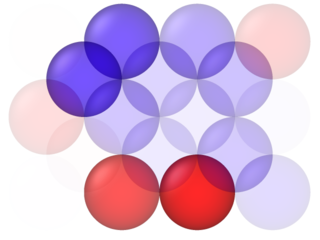}}%
\raisebox{-.5\height}{\includegraphics[width=0.22\textwidth]{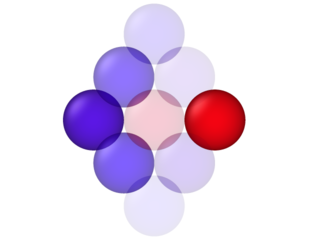}}%
\raisebox{-.5\height}{\includegraphics[width=0.22\textwidth]{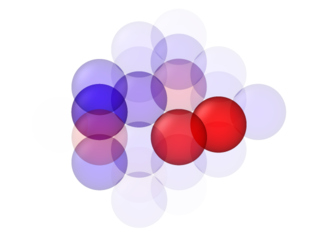}}%
\\%
(2)%
\raisebox{-.5\height}{\includegraphics[width=0.22\textwidth]{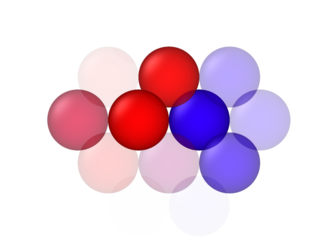}}%
\raisebox{-.5\height}{\includegraphics[width=0.22\textwidth]{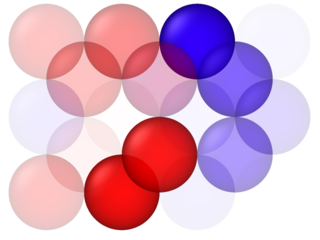}}%
\raisebox{-.5\height}{\includegraphics[width=0.22\textwidth]{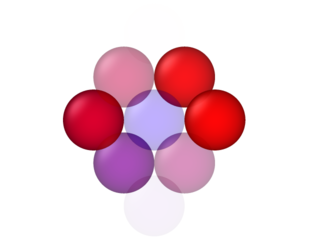}}%
\raisebox{-.5\height}{\includegraphics[width=0.22\textwidth]{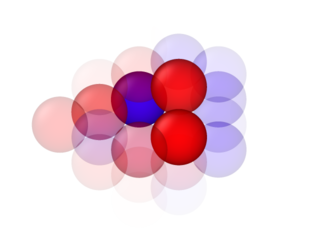}}%
\\%
(3)%
\raisebox{-.5\height}{\includegraphics[width=0.22\textwidth]{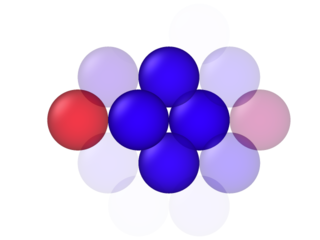}}%
\raisebox{-.5\height}{\includegraphics[width=0.22\textwidth]{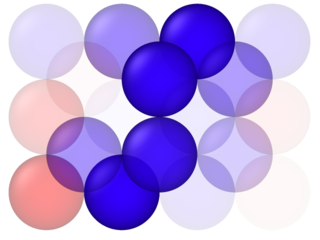}}%
\raisebox{-.5\height}{\includegraphics[width=0.22\textwidth]{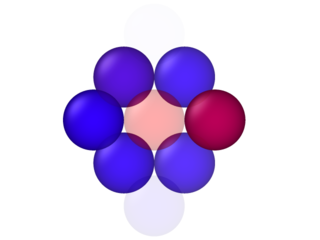}}%
\raisebox{-.5\height}{\includegraphics[width=0.22\textwidth]{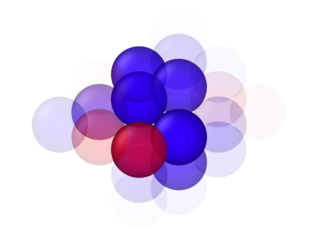}}%
\\%
(4)%
\raisebox{-.5\height}{\includegraphics[width=0.22\textwidth]{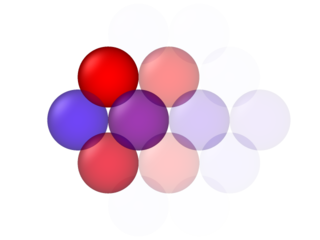}}%
\raisebox{-.5\height}{\includegraphics[width=0.22\textwidth]{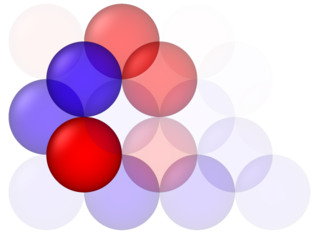}}%
\raisebox{-.5\height}{\includegraphics[width=0.22\textwidth]{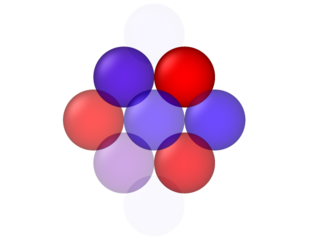}}%
\raisebox{-.5\height}{\includegraphics[width=0.22\textwidth]{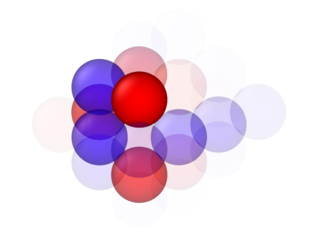}}%
\\%
(5)%
\raisebox{-.5\height}{\includegraphics[width=0.22\textwidth]{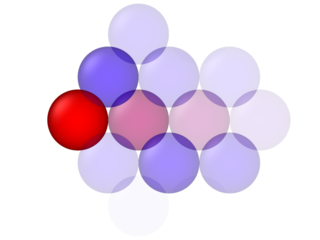}}%
\raisebox{-.5\height}{\includegraphics[width=0.22\textwidth]{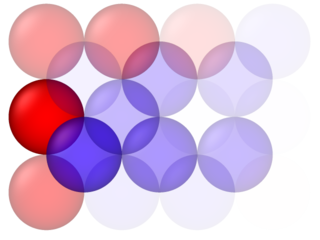}}%
\raisebox{-.5\height}{\includegraphics[width=0.22\textwidth]{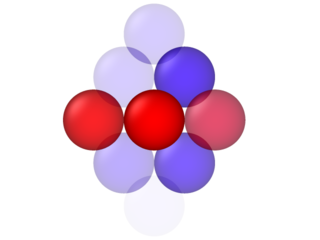}}%
\raisebox{-.5\height}{\includegraphics[width=0.22\textwidth]{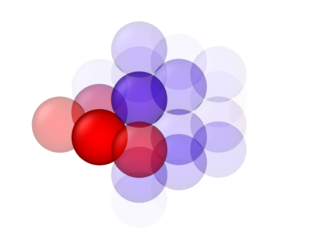}}%
\\%
(6)%
\raisebox{-.5\height}{\includegraphics[width=0.22\textwidth]{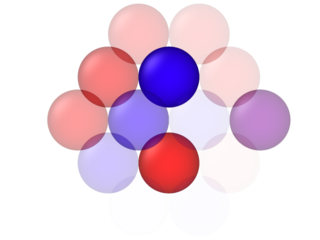}}%
\raisebox{-.5\height}{\includegraphics[width=0.22\textwidth]{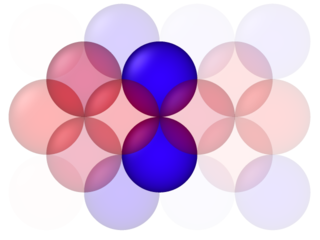}}%
\raisebox{-.5\height}{\includegraphics[width=0.22\textwidth]{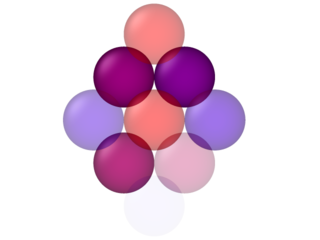}}%
\raisebox{-.5\height}{\includegraphics[width=0.22\textwidth]{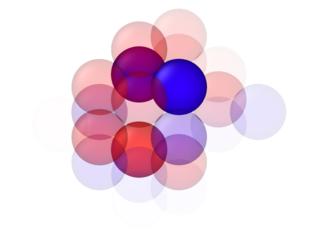}}%
\\%
(7)%
\raisebox{-.5\height}{\includegraphics[width=0.22\textwidth]{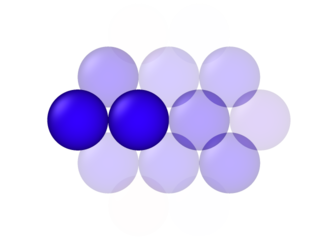}}%
\raisebox{-.5\height}{\includegraphics[width=0.22\textwidth]{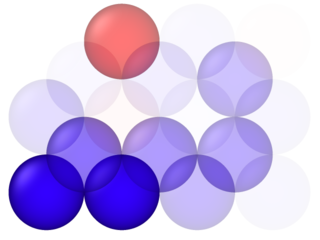}}%
\raisebox{-.5\height}{\includegraphics[width=0.22\textwidth]{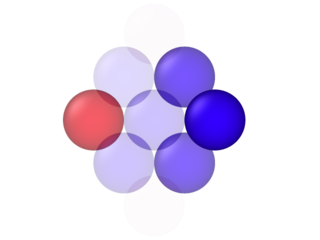}}%
\raisebox{-.5\height}{\includegraphics[width=0.22\textwidth]{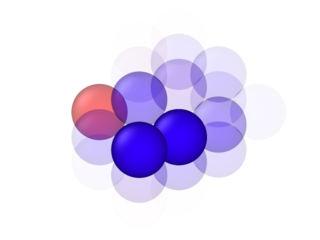}}%
\\%
(8)%
\raisebox{-.5\height}{\includegraphics[width=0.22\textwidth]{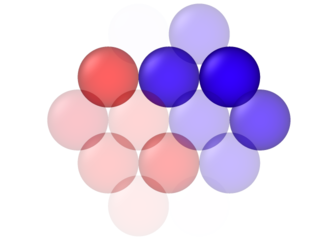}}%
\raisebox{-.5\height}{\includegraphics[width=0.22\textwidth]{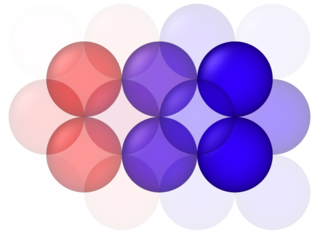}}%
\raisebox{-.5\height}{\includegraphics[width=0.22\textwidth]{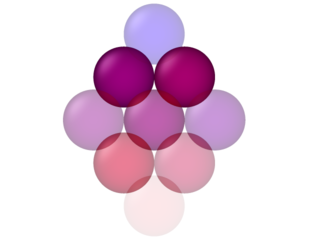}}%
\raisebox{-.5\height}{\includegraphics[width=0.22\textwidth]{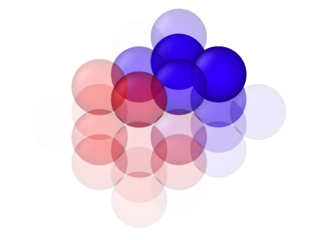}}%
\caption{Prototypes 1--8 of the \hkl{110} RBFN.}%
\label{fig:centroids110a}%
\end{figure}

\begin{figure}%
\raggedleft%
\raisebox{-.5\height}{\includegraphics[width=0.22\textwidth]{side_coordinates.png}}%
\raisebox{-.5\height}{\includegraphics[width=0.22\textwidth]{top_coordinates.png}}%
\raisebox{-.5\height}{\includegraphics[width=0.22\textwidth]{back_coordinates.png}}%
\raisebox{-.5\height}{\includegraphics[width=0.22\textwidth]{ortho_coordinates.png}}%
\\%
(9)%
\raisebox{-.5\height}{\includegraphics[width=0.22\textwidth]{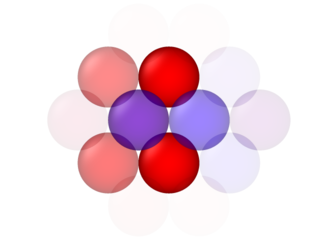}}%
\raisebox{-.5\height}{\includegraphics[width=0.22\textwidth]{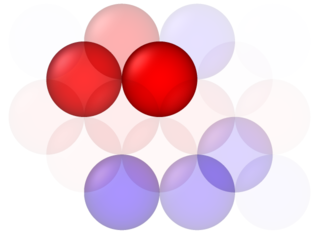}}%
\raisebox{-.5\height}{\includegraphics[width=0.22\textwidth]{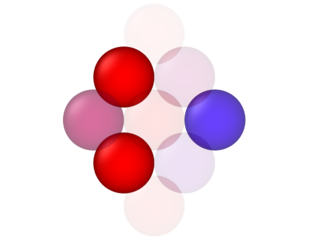}}%
\raisebox{-.5\height}{\includegraphics[width=0.22\textwidth]{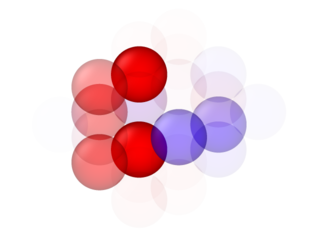}}%
\\%
(10)%
\raisebox{-.5\height}{\includegraphics[width=0.22\textwidth]{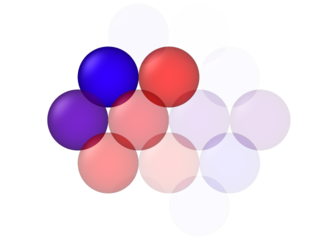}}%
\raisebox{-.5\height}{\includegraphics[width=0.22\textwidth]{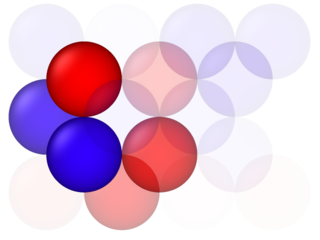}}%
\raisebox{-.5\height}{\includegraphics[width=0.22\textwidth]{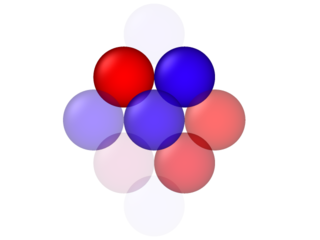}}%
\raisebox{-.5\height}{\includegraphics[width=0.22\textwidth]{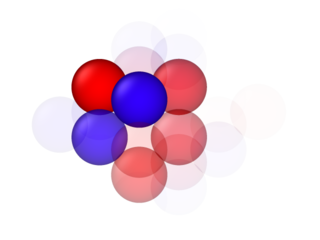}}%
\\%
(11)%
\raisebox{-.5\height}{\includegraphics[width=0.22\textwidth]{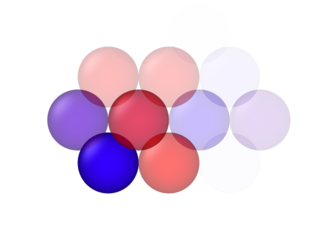}}%
\raisebox{-.5\height}{\includegraphics[width=0.22\textwidth]{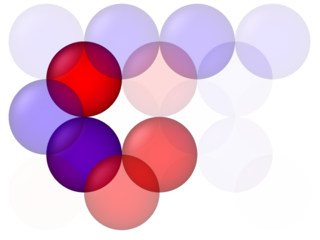}}%
\raisebox{-.5\height}{\includegraphics[width=0.22\textwidth]{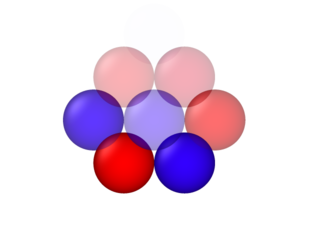}}%
\raisebox{-.5\height}{\includegraphics[width=0.22\textwidth]{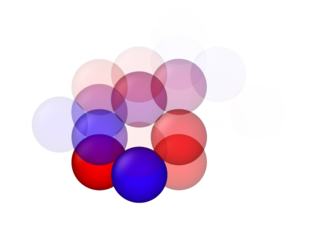}}%
\\%
(12)%
\raisebox{-.5\height}{\includegraphics[width=0.22\textwidth]{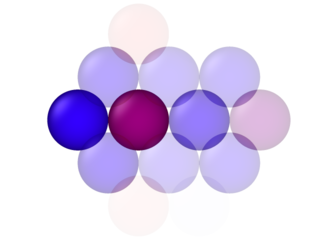}}%
\raisebox{-.5\height}{\includegraphics[width=0.22\textwidth]{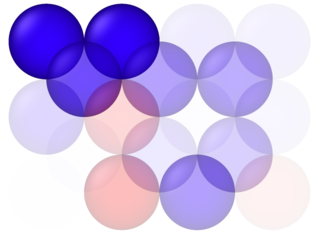}}%
\raisebox{-.5\height}{\includegraphics[width=0.22\textwidth]{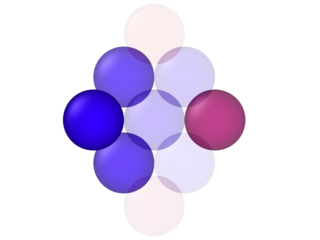}}%
\raisebox{-.5\height}{\includegraphics[width=0.22\textwidth]{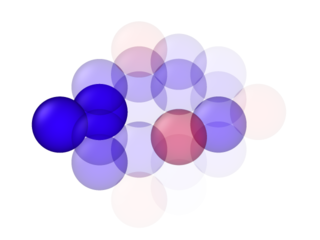}}%
\\%
(13)%
\raisebox{-.5\height}{\includegraphics[width=0.22\textwidth]{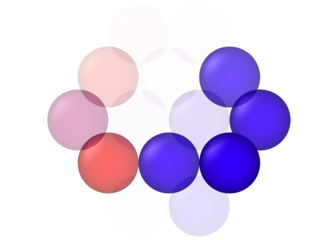}}%
\raisebox{-.5\height}{\includegraphics[width=0.22\textwidth]{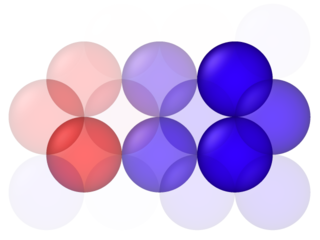}}%
\raisebox{-.5\height}{\includegraphics[width=0.22\textwidth]{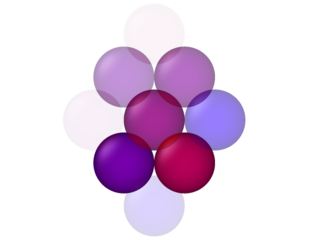}}%
\raisebox{-.5\height}{\includegraphics[width=0.22\textwidth]{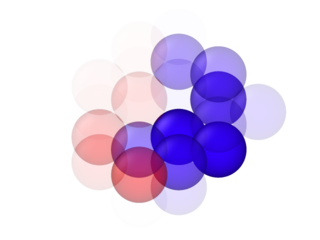}}%
\\%
(14)%
\raisebox{-.5\height}{\includegraphics[width=0.22\textwidth]{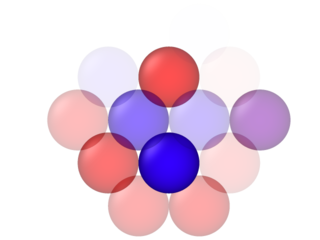}}%
\raisebox{-.5\height}{\includegraphics[width=0.22\textwidth]{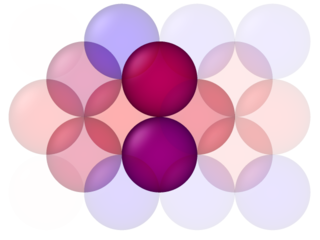}}%
\raisebox{-.5\height}{\includegraphics[width=0.22\textwidth]{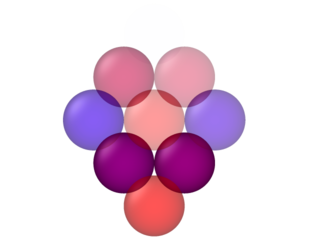}}%
\raisebox{-.5\height}{\includegraphics[width=0.22\textwidth]{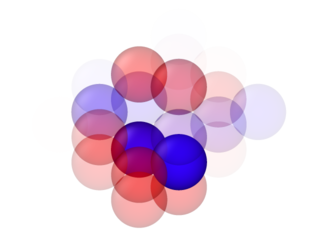}}%
\\%
(15)%
\raisebox{-.5\height}{\includegraphics[width=0.22\textwidth]{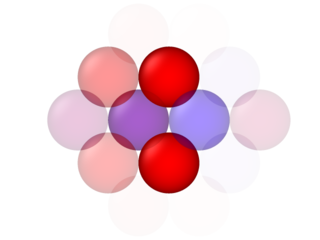}}%
\raisebox{-.5\height}{\includegraphics[width=0.22\textwidth]{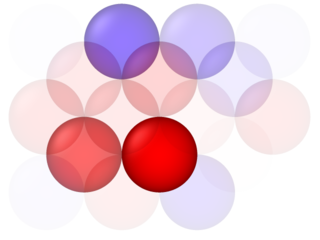}}%
\raisebox{-.5\height}{\includegraphics[width=0.22\textwidth]{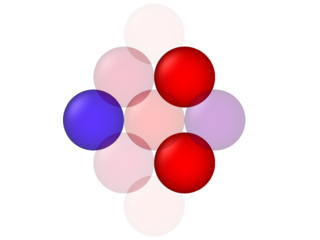}}%
\raisebox{-.5\height}{\includegraphics[width=0.22\textwidth]{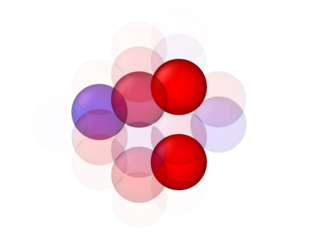}}%
\\%
(16)%
\raisebox{-.5\height}{\includegraphics[width=0.22\textwidth]{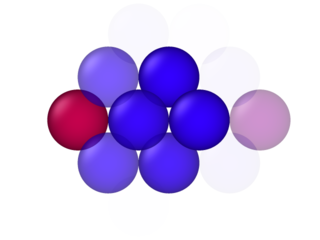}}%
\raisebox{-.5\height}{\includegraphics[width=0.22\textwidth]{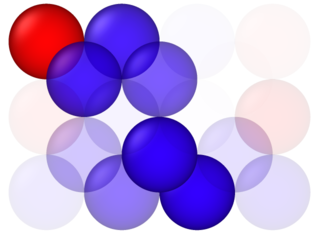}}%
\raisebox{-.5\height}{\includegraphics[width=0.22\textwidth]{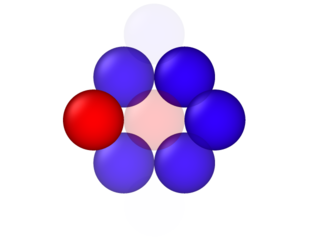}}%
\raisebox{-.5\height}{\includegraphics[width=0.22\textwidth]{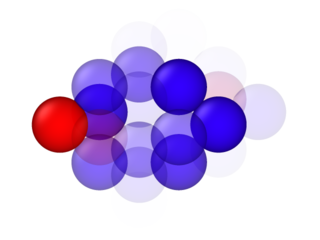}}%
\caption{Prototypes 9--16 of the \hkl{110} RBFN.}%
\label{fig:centroids110b}%
\end{figure}

\begin{figure}%
\raggedleft%
\raisebox{-.5\height}{\includegraphics[width=0.22\textwidth]{side_coordinates.png}}%
\raisebox{-.5\height}{\includegraphics[width=0.22\textwidth]{top_coordinates.png}}%
\raisebox{-.5\height}{\includegraphics[width=0.22\textwidth]{back_coordinates.png}}%
\raisebox{-.5\height}{\includegraphics[width=0.22\textwidth]{ortho_coordinates.png}}%
\\%
(1)%
\raisebox{-.5\height}{\includegraphics[width=0.22\textwidth]{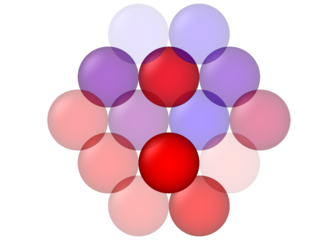}}%
\raisebox{-.5\height}{\includegraphics[width=0.22\textwidth]{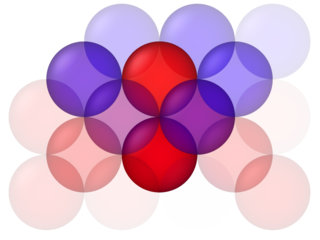}}%
\raisebox{-.5\height}{\includegraphics[width=0.22\textwidth]{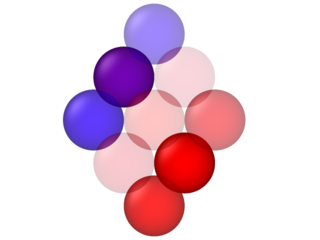}}%
\raisebox{-.5\height}{\includegraphics[width=0.22\textwidth]{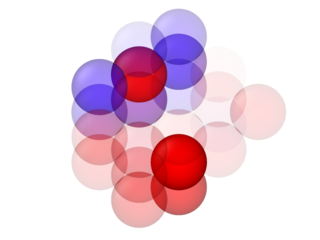}}%
\\%
(2)%
\raisebox{-.5\height}{\includegraphics[width=0.22\textwidth]{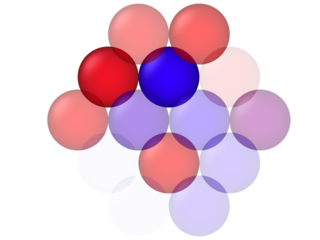}}%
\raisebox{-.5\height}{\includegraphics[width=0.22\textwidth]{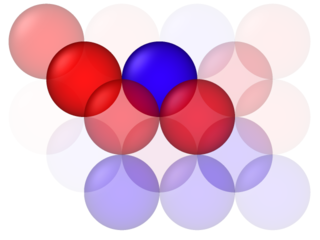}}%
\raisebox{-.5\height}{\includegraphics[width=0.22\textwidth]{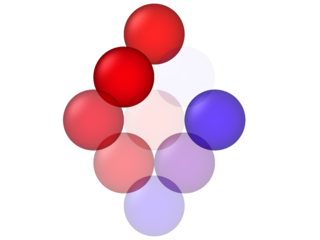}}%
\raisebox{-.5\height}{\includegraphics[width=0.22\textwidth]{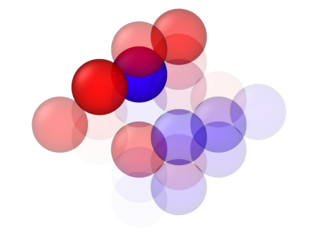}}%
\\%
(3)%
\raisebox{-.5\height}{\includegraphics[width=0.22\textwidth]{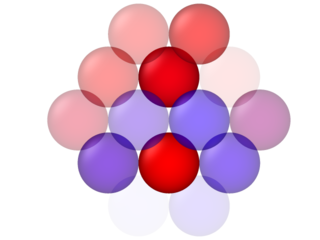}}%
\raisebox{-.5\height}{\includegraphics[width=0.22\textwidth]{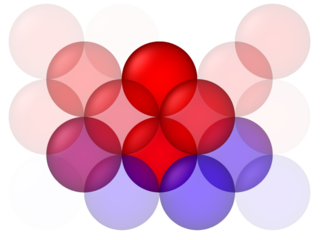}}%
\raisebox{-.5\height}{\includegraphics[width=0.22\textwidth]{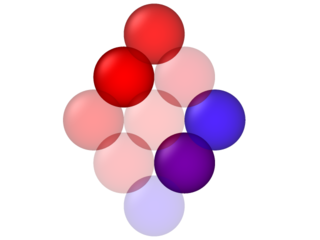}}%
\raisebox{-.5\height}{\includegraphics[width=0.22\textwidth]{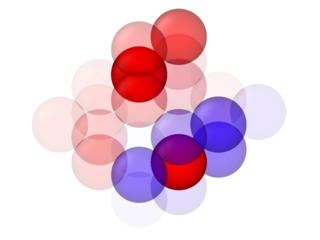}}%
\\%
(4)%
\raisebox{-.5\height}{\includegraphics[width=0.22\textwidth]{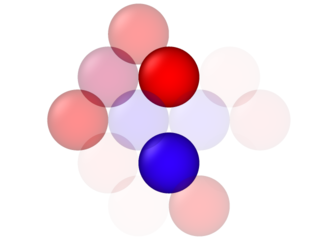}}%
\raisebox{-.5\height}{\includegraphics[width=0.22\textwidth]{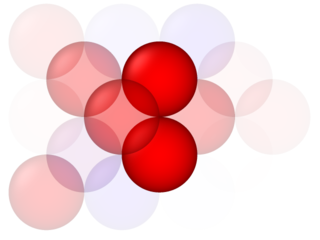}}%
\raisebox{-.5\height}{\includegraphics[width=0.22\textwidth]{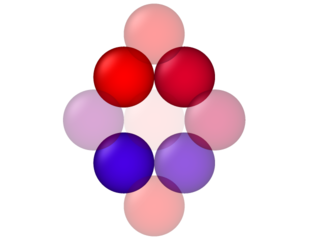}}%
\raisebox{-.5\height}{\includegraphics[width=0.22\textwidth]{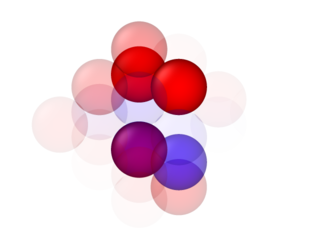}}%
\\%
(5)%
\raisebox{-.5\height}{\includegraphics[width=0.22\textwidth]{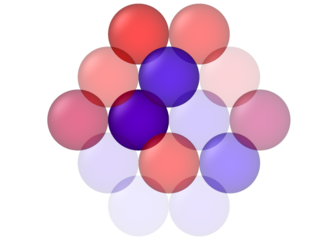}}%
\raisebox{-.5\height}{\includegraphics[width=0.22\textwidth]{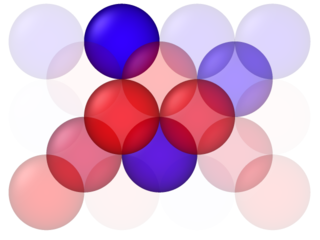}}%
\raisebox{-.5\height}{\includegraphics[width=0.22\textwidth]{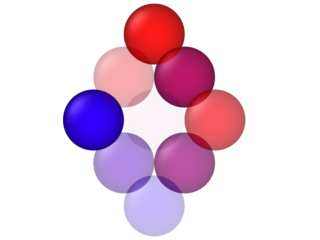}}%
\raisebox{-.5\height}{\includegraphics[width=0.22\textwidth]{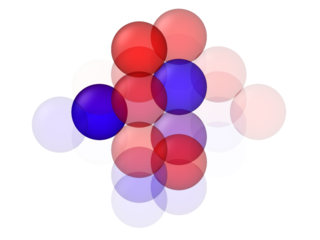}}%
\\%
(6)%
\raisebox{-.5\height}{\includegraphics[width=0.22\textwidth]{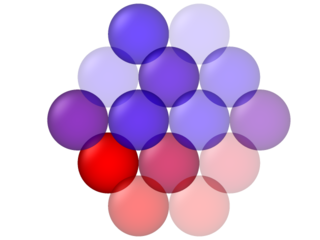}}%
\raisebox{-.5\height}{\includegraphics[width=0.22\textwidth]{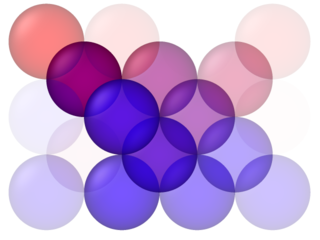}}%
\raisebox{-.5\height}{\includegraphics[width=0.22\textwidth]{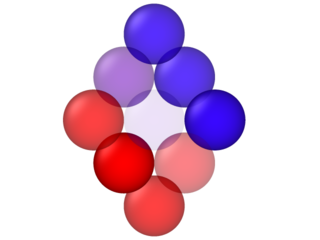}}%
\raisebox{-.5\height}{\includegraphics[width=0.22\textwidth]{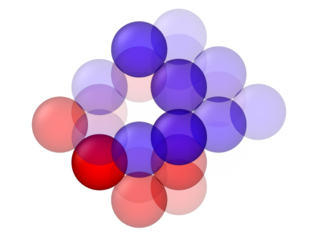}}%
\\%
(7)%
\raisebox{-.5\height}{\includegraphics[width=0.22\textwidth]{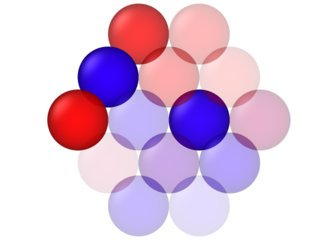}}%
\raisebox{-.5\height}{\includegraphics[width=0.22\textwidth]{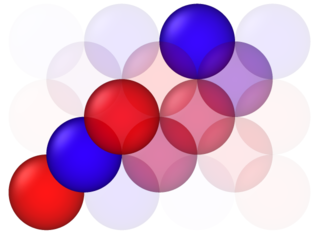}}%
\raisebox{-.5\height}{\includegraphics[width=0.22\textwidth]{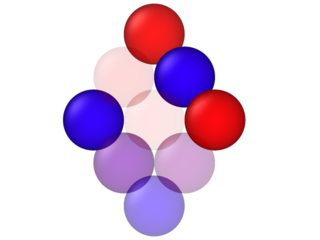}}%
\raisebox{-.5\height}{\includegraphics[width=0.22\textwidth]{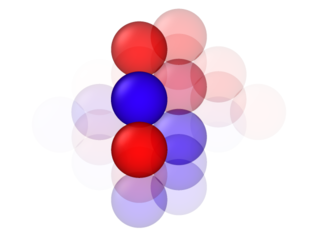}}%
\\%
(8)%
\raisebox{-.5\height}{\includegraphics[width=0.22\textwidth]{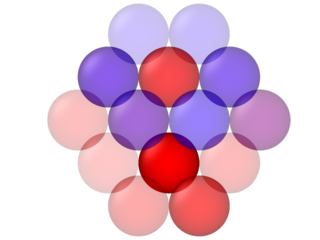}}%
\raisebox{-.5\height}{\includegraphics[width=0.22\textwidth]{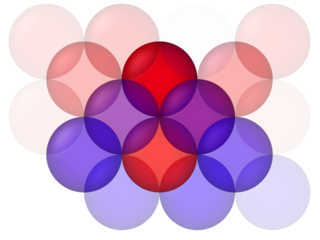}}%
\raisebox{-.5\height}{\includegraphics[width=0.22\textwidth]{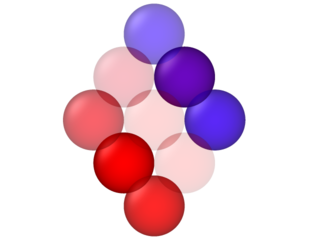}}%
\raisebox{-.5\height}{\includegraphics[width=0.22\textwidth]{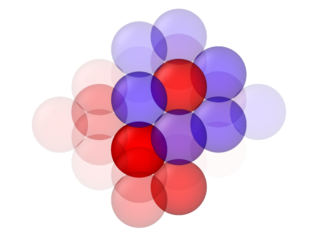}}%
\caption{Prototypes 1--8 of the \hkl{111} RBFN.}%
\label{fig:centroids111a}%
\end{figure}

\begin{figure}%
\raggedleft%
\raisebox{-.5\height}{\includegraphics[width=0.22\textwidth]{side_coordinates.png}}%
\raisebox{-.5\height}{\includegraphics[width=0.22\textwidth]{top_coordinates.png}}%
\raisebox{-.5\height}{\includegraphics[width=0.22\textwidth]{back_coordinates.png}}%
\raisebox{-.5\height}{\includegraphics[width=0.22\textwidth]{ortho_coordinates.png}}%
\\%
(9)%
\raisebox{-.5\height}{\includegraphics[width=0.22\textwidth]{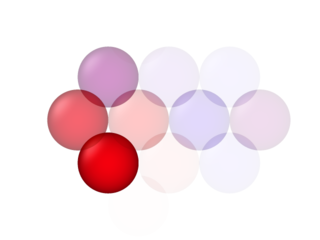}}%
\raisebox{-.5\height}{\includegraphics[width=0.22\textwidth]{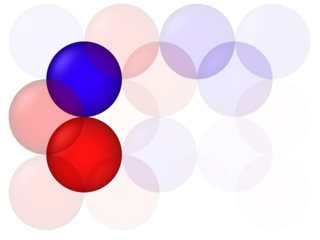}}%
\raisebox{-.5\height}{\includegraphics[width=0.22\textwidth]{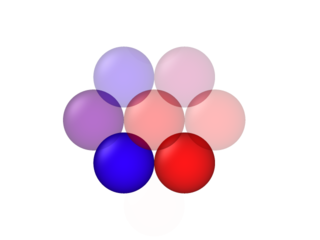}}%
\raisebox{-.5\height}{\includegraphics[width=0.22\textwidth]{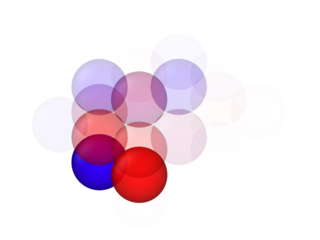}}%
\\%
(10)%
\raisebox{-.5\height}{\includegraphics[width=0.22\textwidth]{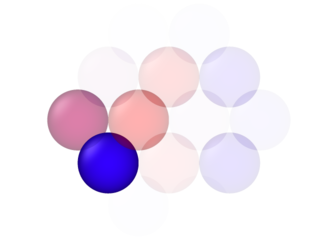}}%
\raisebox{-.5\height}{\includegraphics[width=0.22\textwidth]{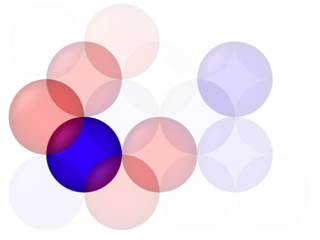}}%
\raisebox{-.5\height}{\includegraphics[width=0.22\textwidth]{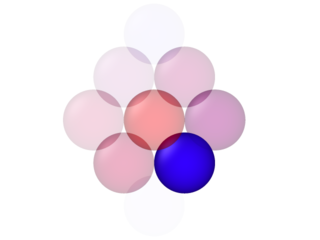}}%
\raisebox{-.5\height}{\includegraphics[width=0.22\textwidth]{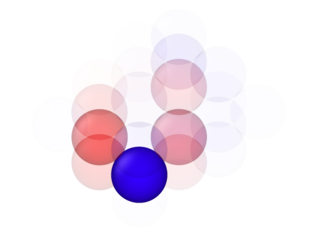}}%
\\%
(11)%
\raisebox{-.5\height}{\includegraphics[width=0.22\textwidth]{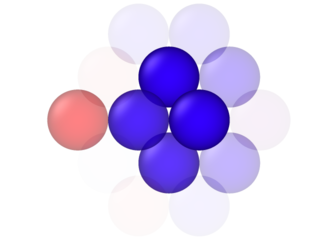}}%
\raisebox{-.5\height}{\includegraphics[width=0.22\textwidth]{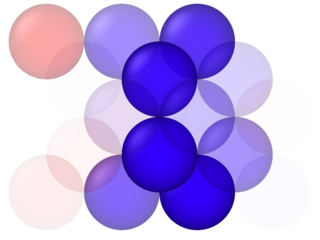}}%
\raisebox{-.5\height}{\includegraphics[width=0.22\textwidth]{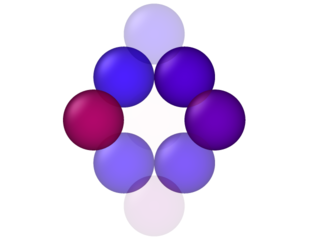}}%
\raisebox{-.5\height}{\includegraphics[width=0.22\textwidth]{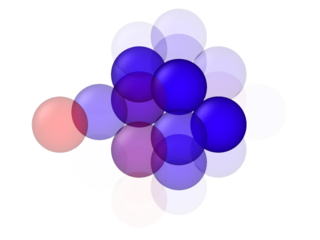}}%
\\%
(12)%
\raisebox{-.5\height}{\includegraphics[width=0.22\textwidth]{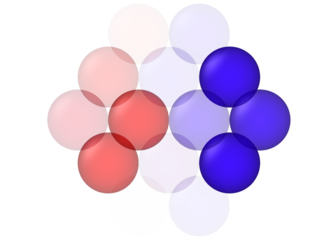}}%
\raisebox{-.5\height}{\includegraphics[width=0.22\textwidth]{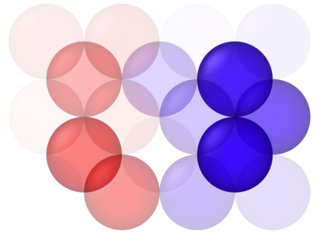}}%
\raisebox{-.5\height}{\includegraphics[width=0.22\textwidth]{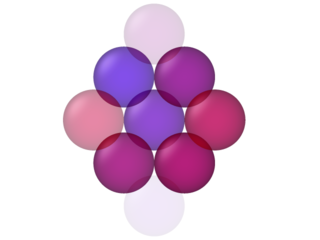}}%
\raisebox{-.5\height}{\includegraphics[width=0.22\textwidth]{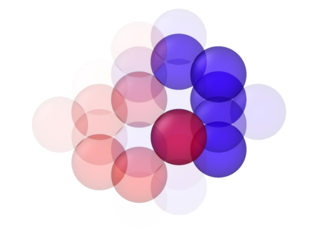}}%
\\%
(13)%
\raisebox{-.5\height}{\includegraphics[width=0.22\textwidth]{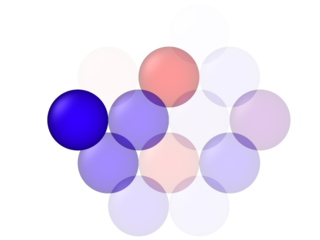}}%
\raisebox{-.5\height}{\includegraphics[width=0.22\textwidth]{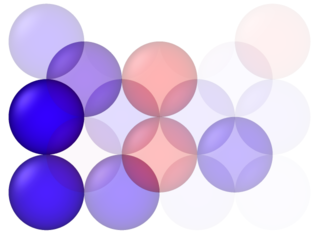}}%
\raisebox{-.5\height}{\includegraphics[width=0.22\textwidth]{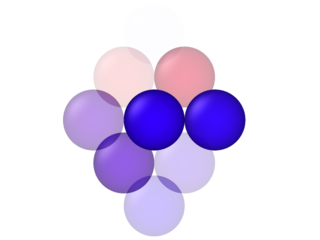}}%
\raisebox{-.5\height}{\includegraphics[width=0.22\textwidth]{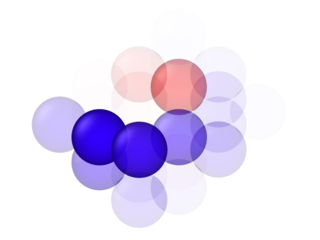}}%
\\%
(14)%
\raisebox{-.5\height}{\includegraphics[width=0.22\textwidth]{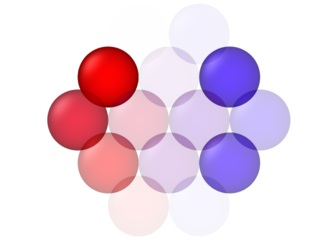}}%
\raisebox{-.5\height}{\includegraphics[width=0.22\textwidth]{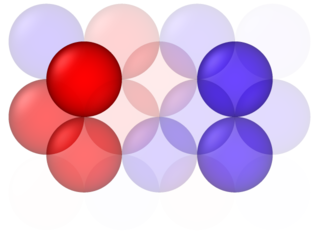}}%
\raisebox{-.5\height}{\includegraphics[width=0.22\textwidth]{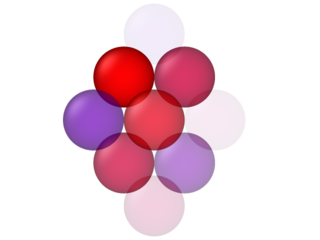}}%
\raisebox{-.5\height}{\includegraphics[width=0.22\textwidth]{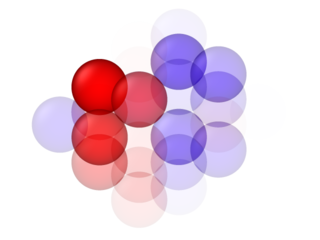}}%
\\%
(15)%
\raisebox{-.5\height}{\includegraphics[width=0.22\textwidth]{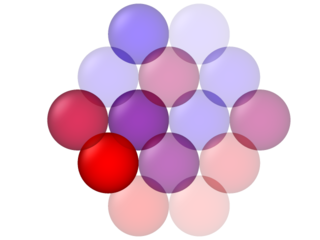}}%
\raisebox{-.5\height}{\includegraphics[width=0.22\textwidth]{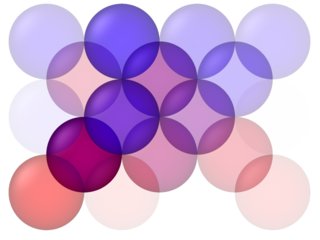}}%
\raisebox{-.5\height}{\includegraphics[width=0.22\textwidth]{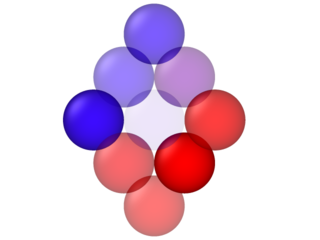}}%
\raisebox{-.5\height}{\includegraphics[width=0.22\textwidth]{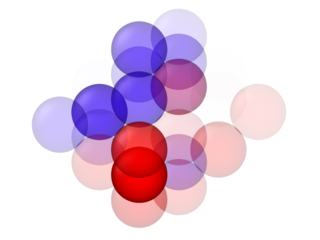}}%
\\%
(16)%
\raisebox{-.5\height}{\includegraphics[width=0.22\textwidth]{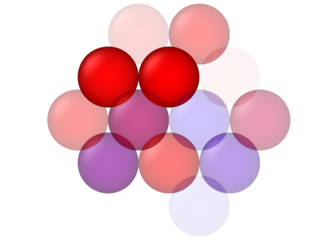}}%
\raisebox{-.5\height}{\includegraphics[width=0.22\textwidth]{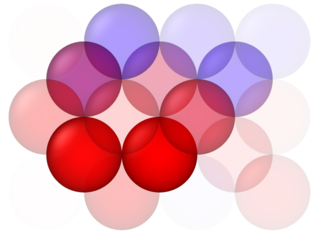}}%
\raisebox{-.5\height}{\includegraphics[width=0.22\textwidth]{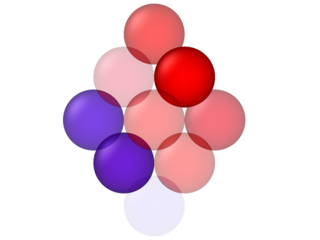}}%
\raisebox{-.5\height}{\includegraphics[width=0.22\textwidth]{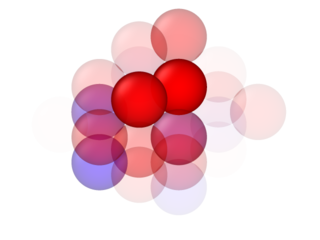}}%
\caption{Prototypes 9--16 of the \hkl{111} RBFN.}%
\label{fig:centroids111b}%
\end{figure}

\end{document}